\documentclass{article}

\usepackage{microtype}
\usepackage{graphicx}
\graphicspath{ {./figures/} } 
\usepackage{subcaption}
\usepackage{booktabs} 
\usepackage{stfloats}
\usepackage{tikz}
\usetikzlibrary{shapes.geometric, arrows.meta, positioning, fit, backgrounds, calc, shadows}

\usepackage{hyperref}

\usepackage[preprint]{icml2026}

\usepackage{amsmath}
\usepackage{amssymb}
\usepackage{mathtools}
\usepackage{amsthm}

\usepackage[capitalize,noabbrev]{cleveref}

\theoremstyle{plain}

\theoremstyle{definition}

\theoremstyle{remark}

\usepackage[disable,textsize=tiny]{todonotes}

\icmltitlerunning{A Dataset is Worth 1 MB}

\begin{document}

\twocolumn[
  \icmltitle{A Dataset is Worth 1 MB}




  \begin{icmlauthorlist}
    \icmlauthor{Elad Kimchi Shoshani}{huji}
    \icmlauthor{Leeyam Gabay}{huji}
    \icmlauthor{Yedid Hoshen}{huji}
  \end{icmlauthorlist}

  \icmlaffiliation{huji}{School of Computer Science and Engineering, The Hebrew University of Jerusalem, Israel}

  \icmlcorrespondingauthor{Elad Kimchi Shoshani}{elad.shoshani@mail.huji.ac.il}

  \icmlkeywords{Machine Learning, ICML, Dataset Distillation, Knowledge Distillation, Data Compression, Pseudo-Labels, Label Distillation, Out-of-Distribution, Dataset Pruning, Efficiency}

  \vskip 0.3in
]



\printAffiliationsAndNotice{}  

\begin{abstract} 
A dataset server must often distribute the same large payload to many clients, incurring massive communication costs. Since clients frequently operate on diverse hardware and software frameworks, transmitting a pre-trained model is often infeasible; instead, agents require raw data to train their own task-specific models locally. While dataset distillation attempts to compress training signals, current methods struggle to scale to high-resolution data and rarely achieve sufficiently small files. In this paper, we propose \textit{Pseudo-Labels as Data} (PLADA), a method that completely eliminates pixel transmission. We assume agents are preloaded with a large, generic, unlabeled \textit{reference} dataset (e.g., ImageNet-1K, ImageNet-21K) and communicate a new task by transmitting only the class labels for specific images. To address the distribution mismatch between the reference and target datasets, we introduce a pruning mechanism that filters the reference dataset to retain only the labels of the most semantically relevant images for the target task. This selection process simultaneously maximizes training efficiency and minimizes transmission payload. Experiments on 10 diverse datasets demonstrate that our approach can transfer task knowledge with a payload of less than 1 MB while retaining high classification accuracy, offering a promising solution for efficient dataset serving.
\end{abstract}

\section{Introduction}

Sending training datasets from a central server to multiple clients is an expensive process, as large datasets must be transmitted repeatedly. This places a heavy burden on dataset servers. Reducing this communication cost is therefore critical. Crucially, sending pre-trained model weights instead of datasets is often insufficient. In many practical scenarios, clients are heterogeneous—ranging from diverse autonomous vehicles to medical devices—and must train models using specific software frameworks (e.g., PyTorch, JAX) or bespoke hardware. Consequently, the server must transmit the training data to allow agents to optimize their own unique models locally. A secondary challenge arises in scenarios where the communication channel is severely bandwidth-constrained. Examples include underwater acoustic links to deep-sea vehicles (up to $\sim$5 kbps) \cite{linkquest_uwm_datasheet,won2012micromodem,annalakshmi2017underwater} or the Titan rover where direct-to-Earth links can be on the order of $\sim$500--800~bps \cite{abelson2005opag,oleson2015titan}. In such cases, transmitting a typical mid-sized (1 GB) dataset would take days to months and can be energetically prohibitive. 
Addressing these scenarios requires methods capable of compressing training datasets by orders of magnitude while minimizing accuracy loss.

A prominent line of research addressing this challenge is dataset distillation. This approach aims to replace a large training dataset with a compact set of synthetic images and labels, such that a model trained on them minimizes regret with respect to the original data. Despite its promise, synthesizing these learning-efficient images is computationally and numerically challenging. While easier on smaller benchmarks like CIFAR-10 \cite{krizhevsky2009learning}, scaling these methods to high-resolution datasets is difficult. The core challenge lies in unrolling optimization steps \cite{cui2023scaling} due to high memory requirements and unstable inner loop optimization. Furthermore, the continuous, full-precision nature of these synthetic pixels often results in file sizes that remain prohibitively large. Finally, dataset distillation struggles with mismatched server and client architectures, although there has been progress on this front.

In this paper, we invert the standard framework of dataset distillation. Rather than synthesizing images while keeping the labels fixed, we synthesize labels while keeping the images fixed. We assume each remote agent is preloaded with a standardized, large, unlabeled set of images, which we term the \textit{reference} dataset. To communicate a new task, we do not transmit pixels; instead, we provide only the class labels for specific images within this reference dataset. Since a label is merely a compact integer index, the transmission payload is drastically reduced. The agent then utilizes its stored reference images and the newly received labels to locally train the target model. In essence, we replace expensive pixel transmission with on-device storage and highly compressed labels.

This approach faces two immediate hurdles: distribution mismatch and efficiency. First, the majority of images in a generic reference dataset are likely semantically unrelated to the target task and can hurt learning. Second, for large reference datasets with many classes, transmitting a label for every image remains bandwidth-intensive. We propose a unified solution to both problems: dataset pruning. We select only a small fraction of reference images for training, ignoring the rest. This ensures that only images semantically related to the target task are used, while simultaneously reducing the transmission cost, as indicating that an image should be ignored requires only a single bit. To achieve this, we introduce a pruning heuristic inspired by out-of-distribution (OOD) detection.

We validate our framework on 10 diverse natural-image datasets and 4 medical (OOD) datasets, utilizing unlabeled ImageNet-1K \cite{deng2009imagenet} and ImageNet-21K \cite{ridnik2021imagenetk} as reference datasets. We demonstrate the ability to transmit the information required to learn a novel task in less than 1 MB, often with only small loss in accuracy. We even find non-trivial accuracy when the target datasets are medical and distributionally distant from the reference set. Qualitative analysis confirms that our selection procedure successfully identifies semantically relevant images, validating the method's effectiveness. Furthermore, we analyze the trade-offs between reference dataset size and transmission payload, and provide ablations on different coding schemes.

\begin{figure}[t]
\begin{center}
\centerline{\includegraphics[width=0.47\textwidth]{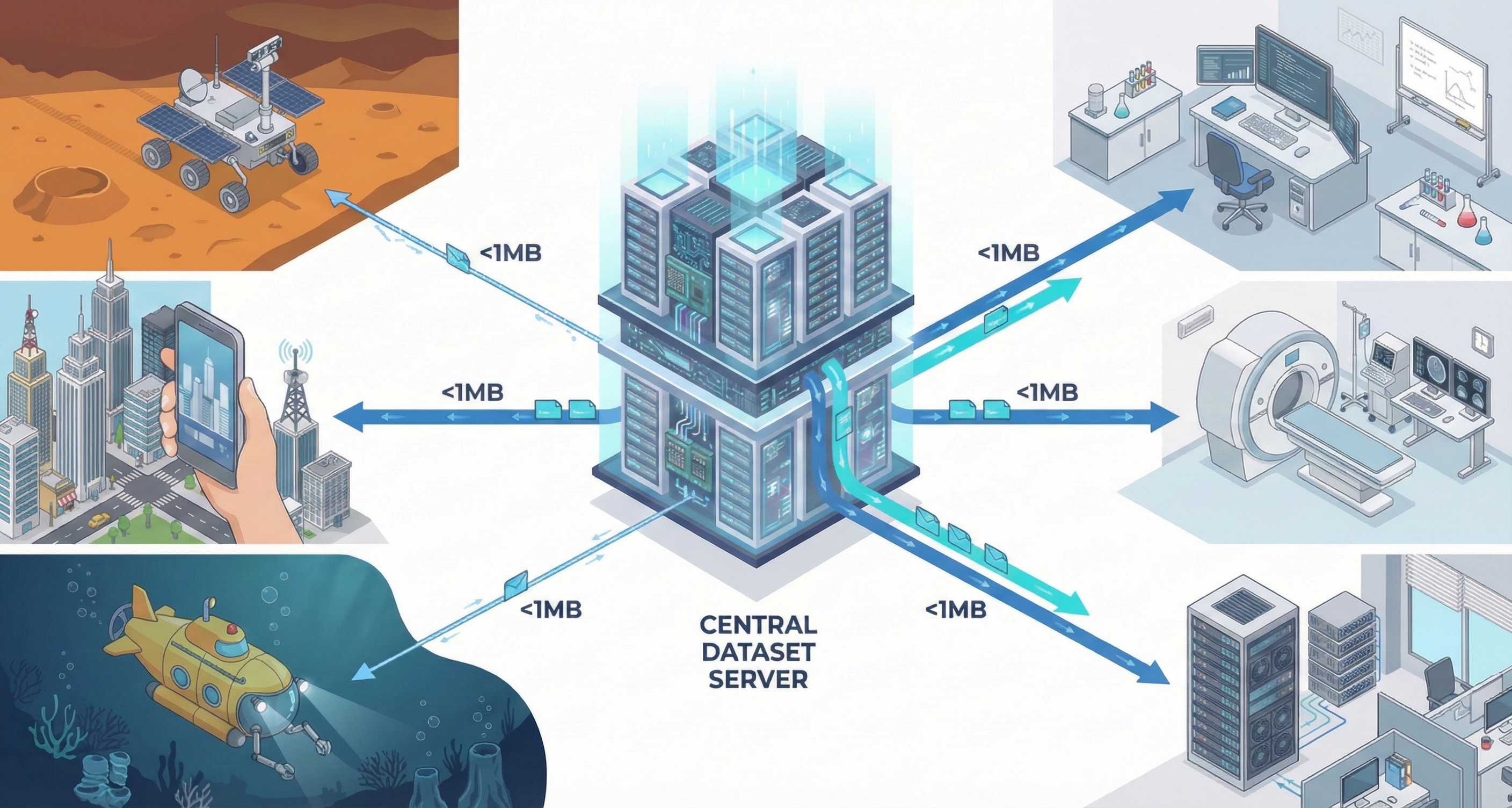}}
\caption{
\textbf{Motivation.} A dataset server transmits the same large dataset many times at massive cost. Our method allows the server to send a compressed payload of less than 1 MB, enabling clients with heterogeneous hardware, even if they have ultra-narrow bandwidth, to train their own models locally.
}
\label{fig:motivation_illustration}
\end{center}
\end{figure}

Our contributions are as follows: 
\begin{enumerate}
\item We propose a new method: \textit{Pseudo-Labels as Data}, which transmits only hard labels while achieving high performance, reducing the transfer payload to a well below one bit per reference image (e.g., 85–206KB at 1\% keep on ImageNet‑21K after Zstd; \cref{tab:compression_results}).
\item We introduce an effective pruning mechanism using Energy-based OOD scores. We show that filtering the reference dataset to just 1\%-10\% of images both improves accuracy and reduces bandwidth costs.
\item We demonstrate that our method achieves high accuracy on diverse classification tasks while transmitting a payload of less than 1 MB.
\end{enumerate}

\section{Related Works}
\textbf{Dataset and Label Distillation.}
Dataset distillation (a.k.a.\ dataset condensation) compresses a full training set into a tiny synthetic set such that training on it approximates training on the original data \cite{wang2018DatasetDistillation,yu2023dataset}. While effective on smaller benchmarks, scaling these methods to high-resolution repositories like ImageNet-21K has historically been limited by exorbitant compute/memory consumption during optimization \cite{zhao2021dataset, cui2023scaling, cazenavette2022dataset, du2023minimizing}. Recent work suggests that the \emph{labels} can be the primary driver of successful distillation—motivating approaches that learn or distill labels rather than synthesizing pixels \cite{sucholutsky2021soft,FlexOn,qin2024LabelWorth1000Images}. PLADA takes this perspective to the extreme: instead of transmitting images, we communicate a task by transmitting only hard pseudo-labels for a fixed, preloaded reference image set.

\textbf{Knowledge Distillation and Pseudo Labels.}
Knowledge distillation trains a \emph{student} model to match the predictions of a \emph{teacher}, typically using soft targets/logits to transfer knowledge across architectures \cite{Hinton2015DistillingTK,nayak2019zero,wang2021knowledge, mansourian2025aKnowledgeDistillation}. When original training data is unavailable, data-free distillation synthesizes inputs \cite{nayak2019zero} or reconstructs them from a trained model \cite{yin2020dreaming}; recent work also frames distillation as an efficient mechanism for faster convergence and improved transfer \cite{he2022knowledge}. Pseudo-labeling and self-training treat a model's high-confidence predictions as supervision, often paired with confidence filtering and meta-learning to improve label quality \cite{lee2013pseudolabel,sohn2020fixmatch,xie2020noisyStudent,pham2021metaPseudoLabels,kage2024reviewPseudoLabeling}. PLADA turns this idea into a communication primitive: a server-side teacher generates hard pseudo-labels on a shared reference dataset, and clients train locally using these \emph{pseudo-labels as data}.

\textbf{OOD Detection and Data Pruning/Selection.}
Deep networks are often overconfident under distribution shift, motivating out-of-distribution (OOD) detection methods based on softmax confidence \cite{HendrycksG17}, temperature/perturbation scoring (ODIN) \cite{liang2018odin}, feature-density scores such as Mahalanobis distance \cite{lee2018mahalanobis}, and energy-based criteria \cite{liu2020energy}. Training-time modifications can also improve confidence separation (e.g., LogitNorm) \cite{wei2022logitnorm,ding2025enhancingELogitNorm}, and recent work further improves distance-based OOD scoring by dynamically calibrating geometry at test time \cite{guo2025ooddcc}. Closely related are data selection and pruning methods that reduce training cost while preserving accuracy \cite{sorscher2022beyond,yang2023datasetpruning}, including approaches that explicitly combine pruning with knowledge distillation to mitigate accuracy loss at high pruning rates \cite{benbaruch2024distillingPruning}. PLADA's pruning stage leverages uncertainty/OOD scores to select semantically relevant reference examples before label transmission, aligning with this literature while operating in a communication-limited dataset-serving setting.

\begin{figure*}[t]
\centering
\centerline{\includegraphics[width=\textwidth]{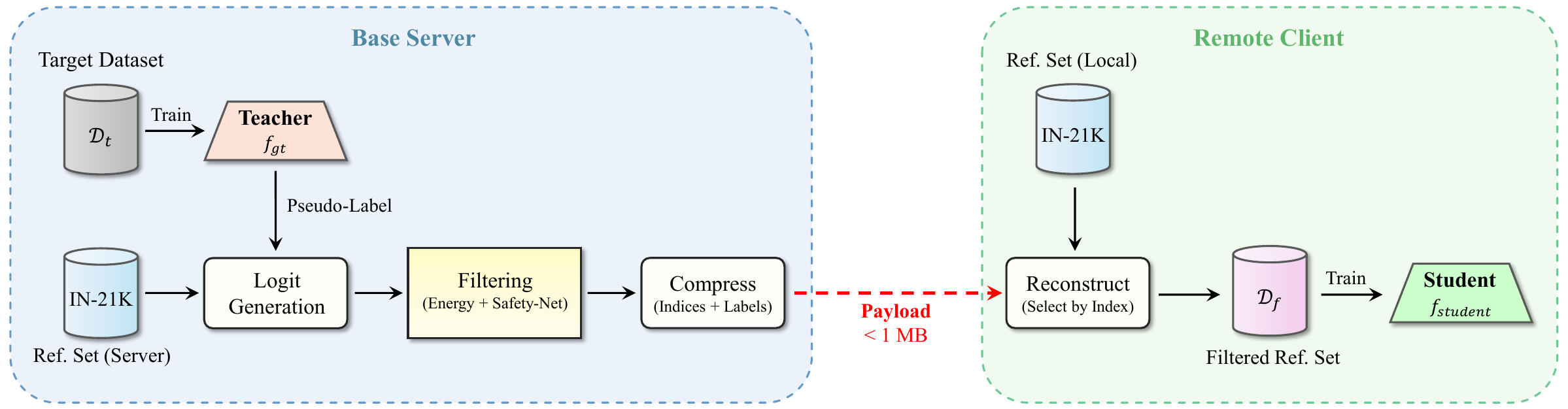}}
\caption{\textbf{The PLADA Pipeline.} The server (left) trains a teacher classifier on the task dataset and distills this task knowledge into hard labels on the reference data. It then filters to the lowest-uncertainty $p\%$ of pseudo-labels and transmits a compressed payload ($<1$ MB). The client (right) reconstructs a virtual dataset using its preloaded reference dataset and the payload to train the student model.}
\label{fig:pipeline}
\end{figure*}

\section{Problem Formulation}
\label{sec:problem_formulation}

In our setting, there is a central server (denoted $A_{s}$) and multiple remote agents (denoted $A_{r}$). We preload all remote agents with the same reference dataset $D_{r}$ containing $n$ unlabeled samples $D_{r}=\{x_{1}, x_{2}, \dots, x_{n}\}$ drawn from a distribution $\mathcal{D}_{r}$. After deployment, when the remote agents are distant from the central server, a new target task arises with distribution $\mathcal{D}_{t}$. Each sample consists of a pair $(x,y)$, where the input is $x$ and the target label is $y$. In this paper, we assume the target label is discrete, making the task a classification problem; we leave the extension to regression for future work. Our objective is to train a classifier $f$ on the remote agent that achieves high accuracy in predicting $y$ given $x$ for $(x,y) \sim \mathcal{D}_{t}$.

To fulfill this task, the server $A_{s}$ transmits a payload $P$ of size $b$ bytes to each remote agent $A_{r}$. We assume the remote agent is capable of training a model given input data. The objective is to maximize the accuracy of classifier $f$ while satisfying the constraint that the transmitted payload does not exceed $b$ bytes.

\section{Method}
\label{sec:method}

\subsection{Overview}
\label{subsec:method:overview}
Our core approach is illustrated in ~\cref{fig:pipeline}. The central server first trains a ground-truth classifier, $f_{gt}$, on the training data from the target distribution $\mathcal{D}_{t}$. It utilizes this classifier to generate pseudo-labels for the reference dataset. The server then transmits these pseudo-labels as the payload to the remote agent. Finally, the remote agent trains a student classifier $f$ on the reference set using the received labels. In \cref{ref_data_pruning}, we present pruning methods to significantly reduce the number of class labels transmitted. In \cref{var_len_coding}, we describe variable-length coding methods that leverage the statistical properties of the labels to further compress the payload size.

\subsection{Efficient Classifier Transfer via Hard Pseudo-Labels}
Transmitting a full dataset to a remote agent requires bandwidth that often exceeds 1 GB. While subsampling datasets (e.g., via coreset selection) can reduce size by 50-80\%, it typically incurs a significant penalty in accuracy. For extreme bandwidth constraints, this reduction is insufficient. Dataset distillation aims to create synthetic images with aggressive compression; however, these methods often result in accuracy loss and still require payloads measuring in megabytes.

Our core premise is that for classification tasks, labels contain far more information per byte than images. However, labels must be associated with images, which are expensive to transmit. To resolve this, we utilize a fixed reference dataset preloaded on each remote agent. To transmit a target task, we send only the pseudo-labels corresponding to the images in this reference dataset. We utilize hard labels rather than soft labels, as storing soft labels requires significantly more memory.

\textbf{Label generation.} Since the reference dataset is generic, many of its images may not correspond to any classes in the target task. We propose a two-step procedure. First, we train a classifier $f_{gt}$ on the training data from the target distribution $\mathcal{D}_{t}$:
\begin{equation}
f_{gt} \leftarrow \operatorname*{arg\,min}_{f} \frac{1}{n_{target}} \sum_{(x,y)\sim\mathcal{D}_{t}} \mathcal{L}_{CE}(f(x),y)
\end{equation}
We then label each image in the reference set using the classifier $f_{gt}$, assigning each image the label corresponding to the maximal logit:
\begin{equation}
l_{i} = \operatorname*{arg\,max}_q f_{gt}(x_{i})[q]
\end{equation}
The server sends a payload consisting of the hard labels for the reference set images:
\begin{equation}
P = [l_{1}, l_{2}, \dots, l_{n}]
\end{equation}
Let $k$ denote the number of target classes. Naively sending the reference set labels requires $n \log_{2} k$ bits. After transmission, the agent trains a classifier based on the locally stored reference images and the received labels:
\begin{equation}
f \leftarrow \operatorname*{arg\,min}_{f^{\prime}} \frac{1}{n} \sum_{i=1}^{n} \mathcal{L}_{CE}(f^{\prime}(x_{i}), l_{i})
\end{equation}
The resulting classifier $f$ on the client serves as the final student model.

\subsection{Reference Dataset Pruning}
\label{ref_data_pruning}
Transmitting a label for every reference image is suboptimal. First, it hurts accuracy: some reference images do not fit any target task classes. For example, an image yielding roughly equal logits for all classes is likely a poor representative for any of them. Forcing such an image into a hard class introduces noise that degrades the target training process. Second, sending a label requires $\log_{2} k$ bits per image, which becomes expensive for large reference sets. Ideally, we should transmit only informative labels.

\begin{figure}[t]
  \vskip 0.2in
  \begin{center}
    \centerline{\includegraphics[width=\columnwidth]{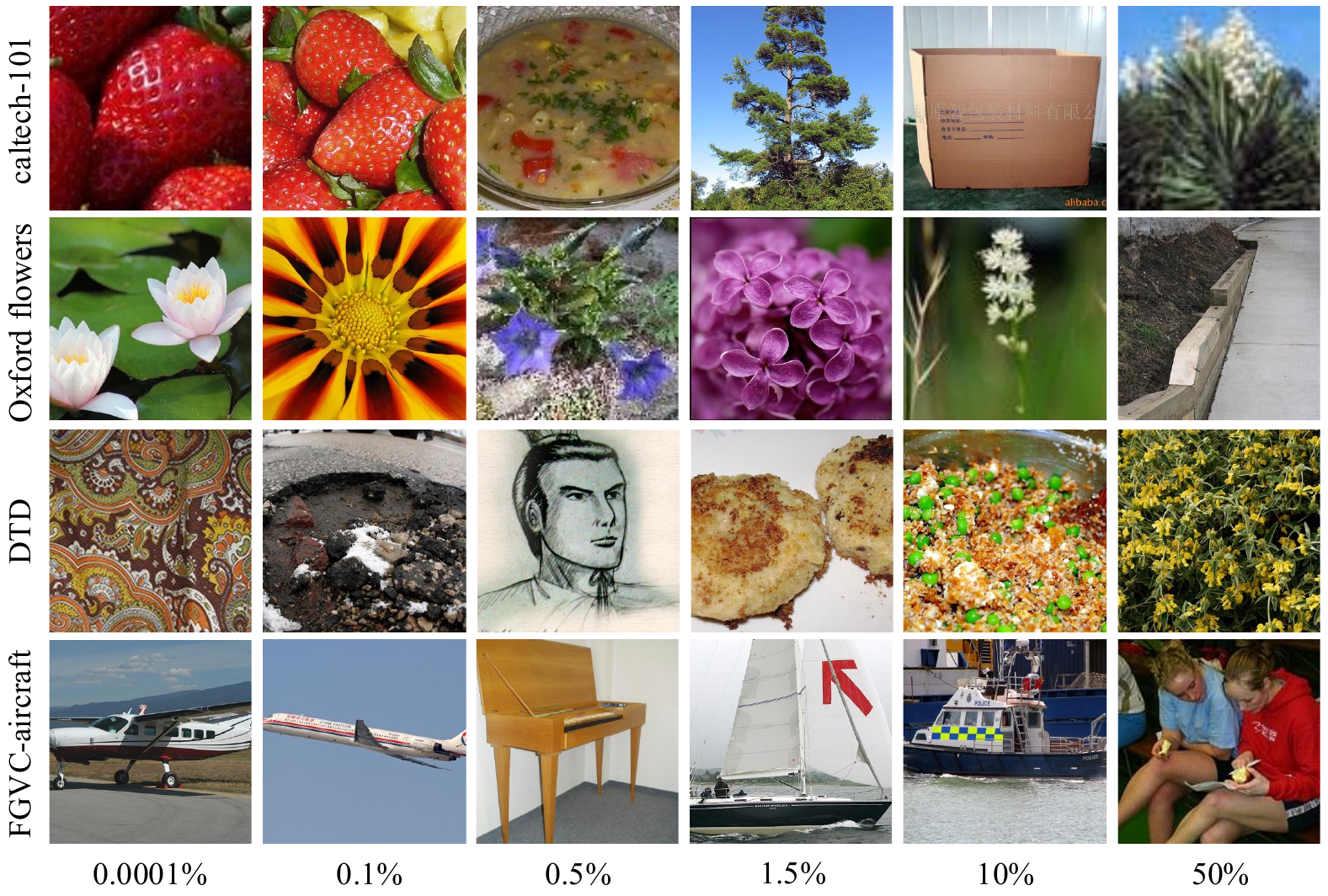}}
    \caption{
    \textbf{Reference set images vs. energy percentile.} High-confidence (low-energy) samples retrieved from ImageNet-21K demonstrate semantic and structural alignment with the target domains. For additional visualizations see Appendix~\ref{sec:app:energy_visualizations}.
    }
    \label{fig:grid_image_by_percentile}
  \end{center}
\end{figure}

\textbf{Selecting informative images.} We draw inspiration from semi-supervised learning, which applies a predictor to a large set of potentially irrelevant images. To isolate relevant samples, these approaches use distribution measures to filter for images where the label certainty is high. Concretely, we retain the top $p\%$ of images based on an uncertainty score (where lower is better). We evaluated several out-of-distribution metrics and found that Logit Energy achieved the best results, with Shannon Entropy performing comparably (see Table \ref{tab:entropy_vs_energy}). We compute energy as:
\begin{equation}
E(x; f_{gt}) = - \log \sum_{j=1}^{k} \exp(f_{gt}(x)[j])
\label{eq:energy_formula}
\end{equation}
In a large reference dataset like ImageNet-21K, typically only a small fraction of images are relevant to a specific downstream task. As shown in \cref{fig:grid_image_by_percentile}, images semantically related to the target dataset often appear only within the top 1\% of lowest energy scores.

Overall, we find that pruning uncertain labels offers three advantages: (i) lower transmission cost, (ii) increased accuracy for the client's target classifier, and (iii) reduced training time due to the smaller dataset size. See \cref{sec:experiments:main_results} for experimental results.

\textbf{Safety-Net Filtering.} While energy-based pruning effectively selects high-confidence samples, it suffers from a significant drawback in high-compression regimes: it disproportionately removes samples from ``harder'' or under-represented classes. When the global retention ratio is low (e.g., 1\%), the filtered dataset is often dominated by a few ``easy'' classes, leading to class collapse and poor student generalization. ~\cref{fig:filtering_bars_plots} illustrates this issue.

To mitigate this, we propose a Safety-Net filtering mechanism. Instead of relying solely on a global energy threshold, we reserve a portion $s$ of the bandwidth budget to ensure that all classes are preserved. We define a class-specific quota $K_{c}$ for each class $c$ based on a power-law weighting of its original size $N_{c}$:
\begin{equation}
K_{c} \propto (N_{c})^{\alpha}
\end{equation}
where $\alpha$ is a balancing hyperparameter.
\begin{itemize}
    \item $\alpha=1$: proportional retention (preserves the original imbalance).
    \item $\alpha=0$: uniform retention (equal quota per class).
    \item $\alpha<0$: tail-favoring retention (weak classes receive larger quotas).
\end{itemize}
We specifically explore negative $\alpha$ values (e.g., $\alpha=-0.2$). This setting intentionally over-samples from smaller or ``weaker'' classes, providing a structural guarantee that tail classes are preserved in the distilled dataset. To construct the final payload, we first fill the Safety-Net quota using the best available samples (lowest energy) per class, and then utilize the remaining budget according to global logit energy, regardless of class membership.

\subsection{Variable-Length Coding}
\label{var_len_coding}
Payload transmission can be optimized with suitable compression. A naive scheme for a payload with $n_{ref}$ reference images and a keep rate of $p\%$ involves sending: (i) one bit per image indicating whether the label was retained, and (ii) $\log_{2} k$ bits for the hard label of each retained image. This results in $b_{raw}$ bits:
\begin{equation}
b_{raw} = n_{ref}(1 + p \log_{2} k)
\end{equation}
For large reference datasets, this overhead is significant. For example, using ImageNet-21K ($\approx 14.2$ million images) as the reference with a 5\% keep rate and 64 classes (6 bits), the payload is approximately 2 MB, the majority of which (1.69 MB) is consumed by the pruning mask (1 bit per reference set image image).

We can mitigate the cost of the pruning mask using Run-Length Encoding (RLE). Instead of storing all bits, we store the distance between consecutive kept indices. For low keep rates ($p \ll 1$), this exploits sparsity effectively, reducing the average cost per selected item significantly compared to a dense bitmap.

\begin{figure}[t]
  \vskip 0.2in
  \begin{center}
    \centerline{\includegraphics[width=\columnwidth]{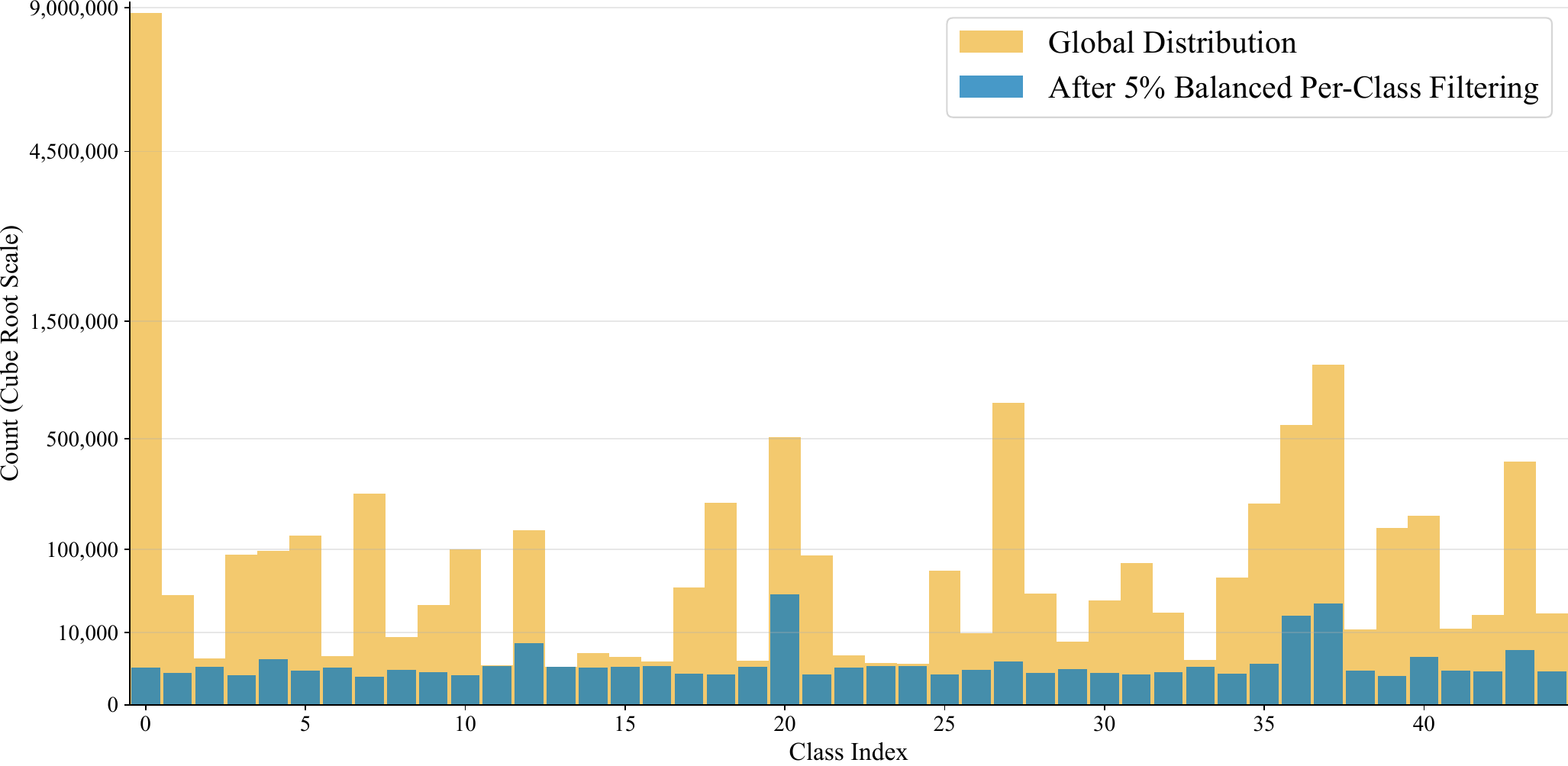}}
    \caption{
        \textbf{Class distribution of the RESISC45 pseudo hard-labels}, before and after filtering using safety-net. The yellow bars show the original global distribution, which is heavily imbalanced - RESISC45 has images extracted using Google Earth, out of which class 0 is \textit{airplane}. Standard global filtering would eliminate some of the tail classes entirely. The blue bars demonstrate our Safety-Net Filtering (keeping 5\%, $\alpha=-0.2$), which effectively preserves a representation of under-represented classes even under extreme compression. Note that the Y-axis uses a cube-root scale to visually accommodate the large magnitude differences between the `strong` and `weak` classes.
    }
    \label{fig:filtering_bars_plots}
  \end{center}
\end{figure}

Furthermore, we exploit the statistical distribution of classes. Instead of using a fixed $\log_{2} k$ bits per label, we employ variable-length encoding so that frequent classes are assigned shorter codes. Huffman coding is a classical method leveraging this principle. We illustrate the class distribution in Fig.~\ref{fig:filtering_bars_plots}.

While these classical concepts highlight the sources of redundancy, modern implementations offer superior performance. In our experiments, we utilize Zstd \cite{collet2021rfc, zstdgithub}, a modern state-of-the-art compression library, to compress the final pseudo-label payloads.

\section{Experiments}
\label{sec:experiments}

In this section, we evaluate the proposed PLADA framework. We assess its ability to transfer task knowledge under extreme bandwidth constraints, analyze its robustness to out-of-distribution (OOD) tasks, and validate the efficacy of the Safety-Net filtering mechanism.

\begin{table*}[t]
\centering
\caption{
\textbf{Keep rate evaluations.} Student accuracy when trained on the top-$p$ images of the reference set, according to logit Energy. 
We achieve accurate classification on target tasks using only a fraction of the reference set.
}
\label{tab:main_results}
\vskip 0.15in
\begin{small}
\setlength{\tabcolsep}{1.4pt}
\begin{sc}
\begin{tabular}{l @{\hspace{6pt}} ccccc c @{\hspace{8pt}} ccccc c}
\toprule
\abovespace
& \multicolumn{6}{c}{ImageNet-21K (14.2M images)} & \multicolumn{6}{c}{ImageNet-1K (1.2M images)} \\
\cmidrule(lr){2-7} \cmidrule(lr){8-13}
Dataset & 1\% & 5\% & 10\% & 25\% & 50\% & 100\%* & 1\% & 5\% & 10\% & 25\% & 50\% & 100\%* \\
\midrule
\abovespace
Caltech-101   & 79.84\% & 88.94\% & 90.21\% & 90.73\% & 92.45\% & \textbf{92.74\%} & 66.59\% & 77.13\% & 82.49\% & 86.35\% & 86.52\% & \textbf{87.50\%} \\
CIFAR-10      & 63.31\% & 85.31\% & 88.12\% & 91.68\% & \textbf{92.62\%} & 86.13\% & 53.75\% & 72.02\% & 74.83\% & 79.90\% & 84.96\% & \textbf{87.66\%} \\
CUB-200       & \textbf{82.49\%} & 82.36\% & 82.44\% & 81.34\% & 81.09\% & 74.94\% & 22.94\% & 45.89\% & 51.40\% & \textbf{56.19\%} & 55.60\% & 52.97\% \\
DTD           & 66.65\% & 70.16\% & 70.69\% & \textbf{70.80\%} & 68.83\% & 68.14\% & 52.45\% & 58.99\% & 60.48\% & 60.90\% & 61.97\% & \textbf{62.29\%} \\
FGVC-Aircraft & \textbf{53.62\%} & 45.51\% & 46.41\% & 45.87\% & 43.53\% & 32.16\% & 20.04\% & 23.91\% & 26.58\% & 29.16\% & \textbf{30.18\%} & 29.40\% \\
Food-101      & 75.50\% & 76.18\% & \textbf{76.91\%} & 76.72\% & 75.27\% & 73.43\% & 37.66\% & 50.82\% & 52.90\% & 56.01\% & 56.95\% & \textbf{57.60\%} \\
Oxford-Flowers& 96.93\% & 98.41\% & \textbf{98.50\%} & 98.19\% & 98.06\% & 96.76\% & 62.29\% & 75.65\% & 75.12\% & \textbf{75.80\%} & 73.38\% & 68.66\% \\
Oxford-IIIT-Pet   & 90.95\% & \textbf{91.03\%} & 90.81\% & 90.05\% & 89.48\% & 87.74\% & 83.21\% & 88.66\% & 88.93\% & 89.26\% & \textbf{89.40\%} & 88.69\% \\
Places365     & 23.39\% & 34.89\% & 40.05\% & 45.82\% & \textbf{48.66\%} & 46.95\% & 16.97\% & 26.85\% & 31.96\% & 38.36\% & 41.63\% & \textbf{43.69\%} \\
RESISC45      & 58.16\% & 67.81\% & 74.37\% & \textbf{80.62\%} & 76.79\% & 31.02\% & 30.03\% & 50.44\% & 57.67\% & 63.60\% & 71.68\% & \textbf{78.73\%} \\
\bottomrule
\end{tabular}
\end{sc}
\begin{flushleft}
\scriptsize
* Indicates no filtering (full reference set used).\\
$^{\dagger}$ \textbf{Teacher Accuracies:} Caltech-101 (98.39\%), CIFAR-10 (98.15\%), CUB-200 (97.71\%), DTD (77.50\%), FGVC-Aircraft (86.53\%), Food-101 (90.02\%), Oxford-Flowers (99.04\%), Oxford-Pets (93.40\%), Places365 (55.45\%), RESISC45 (96.84\%).
\end{flushleft}
\end{small}
\end{table*}

\subsection{Experimental Setup}

\paragraph{Datasets and Benchmarks.} 
We evaluate our method on 14 diverse classification datasets, categorized by domain to test generalization across varying granularities:
\begin{itemize}
    \item \textit{Coarse-grained Objects:} Caltech-101~\cite{li_andreeto_ranzato_perona_2022}, CIFAR-10~\cite{krizhevsky2009learning}, and Places365~\cite{zhou2017places}.
    \item \textit{Fine-grained Classification:} CUB-200-2011~\cite{wah_branson_welinder_perona_belongie_2022}, DTD (Textures)~\cite{cimpoi2014describing}, FGVC-Aircraft~\cite{maji2013fine}, Food-101~\cite{bossard14}, Oxford-Flowers-102~\cite{Nilsback08}, Oxford-IIIT Pet~\cite{parkhi12a}, and RESISC45~\cite{Cheng_2017}.
    \item \textit{Medical (OOD Stress Test):} To test the limits of our approach on data distributionally disjoint from ImageNet, we utilize BloodMNIST, DermaMNIST, RetinaMNIST, and NCT-CRC-HE-100K \cite{medmnistv1,medmnistv2, ignatov2024nct-crc-he-100k}.
\end{itemize}

\noindent\textit{Data Leakage Verification:} We rigorously verified (see App.~\ref{sec:app:intersection}) that there is zero or statistically negligible intersection ($<1\%$) between the target test sets and the ImageNet reference datasets, ensuring that student performance is not a result of memorization.

\paragraph{Baselines.} 
We compare PLADA against three transmission strategies:
\begin{enumerate}
    \item \textit{Random Subset:} Transmitting a balanced random subset of raw target images.
    \item \textit{Coreset Selection:} Selecting representative images via K-Center ~\cite{sener2017kcoreset,moser2025coreset} from each target class.
    \item \textit{Dataset Distillation (DD):} Comparing against state-of-the-art distillation methods where available.
\end{enumerate}
For a fair comparison of information density, all baseline image payloads are compressed using JPEG ($Q=30$) and measured after applying Zstandard (Zstd) compression (level 19).

\paragraph{Implementation Details.} As the teacher model, we use a \texttt{ConvNeXt-V2-Tiny}~\cite{woo2023convnext} pre-trained on ImageNet-21K and fine-tuned on the target data. The remote agent trains a \texttt{ResNet-18}~\cite{he2016deep} initialized with pre-trained weights. We train for 5 epochs (when using the ImageNet-21K reference set) or 30 epochs (ImageNet-1K) using the AdamW optimizer~\cite{loshchilov2018decoupled} ($lr=10^{-3}$, cosine schedule). All experiments were conducted on a single NVIDIA A5000 GPU.

\begin{table*}[ht]
  \caption{
  \textbf{Baseline comparison.} We compare student accuracy across 10 benchmarks. Our method, \textbf{PLADA} (using 1\% keep ratio with possible Safety-Net filtering on ImageNet-21K, see \cref{tab:safetynet_results}), outperforms data-transmission baselines—including random sampling, K-Center coresets, and Dataset Distillation (DD)—in both accuracy and payload size. Notably, PLADA achieves superior task recovery while requiring a payload significantly smaller than even the aggressive 100-image JPEG-compressed subsets.
  }
  \label{tab:baselines}
  \begin{center}
    \begin{small}
      \begin{sc}
        \begin{tabular}{lcccccc}
          \toprule
          \abovespace
        & & \multicolumn{2}{c}{Using 100 Images} & \multicolumn{2}{c}{Using 500 Images} & \\
        \cmidrule(lr){3-4} \cmidrule(lr){5-6}
          Dataset & Ours (p=1\%) & Random & K-Centers & Random & K-Centers & DD$^{\dagger}$  \\
          \midrule
            Caltech-101   & \textbf{86.69\%} & 32.78\% & 34.16\% & 47.98\% & 51.04\% & --  \\
            CIFAR-10      & \textbf{76.75\%} & 28.66\% & 19.33\% & 31.29\% & 27.20\% & 73.2\%  \\
            CUB-200       & \textbf{82.49\%} & 4.58\%  & 3.69\%  & 9.67\%  & 7.55\%  & 16.2\%  \\
            DTD           & \textbf{68.09\%} & 19.04\% & 14.73\% & 36.49\% & 28.35\% & --  \\
            FGVC-Aircraft & \textbf{53.62\%} & 2.76\%  & 2.10\%  & 4.62\%  & 4.59\%  & --  \\
            Food-101      & \textbf{75.50\%} & 3.95\%  & 3.20\%  & 10.26\% & 5.89\%  & 77.6\%  \\
            Oxford-Flowers & \textbf{97.53\%} & 36.39\% & 33.74\% & 34.20\% & 25.78\% & 71.1\%  \\
            Oxford-IIIT-Pet & \textbf{90.98\%} & 11.97\% & 15.91\% & 61.60\% & 53.61\% & --  \\
            Places365     & \textbf{31.59\%} & 1.17\%  & --      & 3.21\%  & 2.91\%      & --  \\
            RESISC45      & \textbf{75.65\%} & 20.81\% & 11.16\% & 29.98\% & 19.57\% & --  \\
          \midrule
          Size* (KB) & \textbf{147.3}$\pm$13.2 & 356.4$\pm$27.8 & 376.9$\pm$30.5 & 1818.0$\pm$126.9 & 1907.7$\pm$136.4 & -- \\
          \bottomrule
        \end{tabular}
      \end{sc}
      \begin{flushleft}
      *Reported payload sizes are mean$\pm$SEM. Baseline payloads are compressed using Zstandard (level 19). \\
      $^{\dagger}$Dataset-Distillation (DD) results: CIFAR-10 \cite{moser2025unlocking}, CUB-200 \cite{shul2025distilling}, Food-101 \cite{hu2025focusdd}, Oxford-Flowers \cite{hu2025focusdd}. 
\end{flushleft}
    \end{small}
  \end{center}
  \vskip -0.1in
\end{table*}


\begin{table}[ht]
\caption{
\textbf{Impact of safety-net filtering} on Student Accuracy. All 1\% subsets are sampled from ImageNet-21K. We compare the default lowest-energy filtering and its counterpart (highest-energy filtering)  against Safety-Net variants. Low-energy samples (Vanilla) outperform high-energy ones. Safety-Net often further improves accuracy by preventing class collapse, with the same payload budget. The difference is the highest for RESISC45 (cf. \cref{fig:filtering_bars_plots}). Additional results are reported in Table ~\ref{tab:safetynet_imagenet_21k_vs_1k}.
}
\label{tab:safetynet_results}
\begin{center}
\begin{small}
\begin{sc}
\setlength{\tabcolsep}{1.6pt} 
\begin{tabular}{lcccc}
\toprule
Dataset & \begin{tabular}{@{}c@{}}1\% \\ Vanil.\end{tabular} & \begin{tabular}{@{}c@{}}1\% \\ Oppos.\end{tabular} & \begin{tabular}{@{}c@{}}1\%+Safe \\ ($\alpha=0.5$)\end{tabular} & \begin{tabular}{@{}c@{}}1\%+Safe \\ ($\alpha=-0.2$)\end{tabular} \\
\midrule
Caltech-101    & 79.84\% & 74.42\% & \textbf{86.69\%} & 86.29\% \\
CIFAR-10       & 63.31\% & 54.80\% & \textbf{76.75\%} & 74.62\% \\
CUB-200        & \textbf{82.49\%} & 6.19\%  & 81.21\% & 80.53\% \\
DTD            & 66.65\% & 26.12\% & 66.70\% & \textbf{68.09\%} \\
FGVC-Aircraft  & \textbf{53.62\%} & 3.45\%  & 43.23\% & 44.58\% \\
Food-101       & \textbf{75.50\%} & 6.28\%  & 70.91\% & 71.66\% \\
Oxford-Flowers & 96.93\% & 9.95\%  & \textbf{97.53\%} & 97.35\% \\
Oxford-IIIT-Pet    & 90.95\% & 18.67\% & \textbf{90.98\%} & 90.87\% \\
Places365      & 23.39\% & 18.04\% & 30.26\% & \textbf{31.59\%} \\
RESISC45       & 58.16\% & 2.06\%  & 72.81\% & \textbf{75.65\%} \\
\bottomrule
\end{tabular}
\end{sc}
\end{small}
\end{center}
\vskip -0.1in
\end{table}

\begin{table}[t]
\centering
\caption{\textbf{Payload size analysis.} We compare across keep ratios ($p$) and compression schemes. Ranges represent the minimum and maximum sizes observed across all 10 different datasets, and across different filtering options (with and without safety-net), with ImageNet-21K as the reference dataset. \textbf{Raw} indicates uncompressed fixed-width binary storage. \textbf{Zstd} represents the final compressed payload size using differential encoding and Zstandard compression (level 19). Full compression experiments results are provided in Appendix~\ref{sec:app:compression}.
}
\label{tab:compression_results}
\resizebox{\columnwidth}{!}{%
\begin{tabular}{lcccc}
\toprule
$p$ & Raw Size & Huffman & Zstd \\ 
\midrule
0.5\%  & 0.41--1.83 MB   & 77--305 KB     & \textbf{45--109 KB}   \\
1\%    & 0.81--1.96 MB   & 151--396 KB    & \textbf{85--206 KB}   \\
5\%    & 3.05 MB         & 0.57--1.10 MB  & \textbf{0.40--0.88 MB} \\
10\%   & 4.40--8.12 MB   & 0.88--1.95 MB  & \textbf{0.67--1.58 MB} \\
25\%   & 8.46 MB         & 1.65--4.34 MB  & \textbf{1.21--3.47 MB} \\
50\%   & 15.23--40.62 MB & 2.49--7.88 MB  & \textbf{1.87--6.42 MB} \\
100\%  & 27.08 MB        & 2.29--12.83 MB & \textbf{1.77--10.50 MB} \\
\bottomrule
\end{tabular}%
}
\end{table}

\subsection{Main Results}
\label{sec:experiments:main_results}

\paragraph{Accuracy vs. bandwidth efficiency.}
Table~\ref{tab:main_results} summarizes student accuracy using ImageNet-21K and ImageNet-1K as reference sets. The results validate our core premise: highly accurate task transfer is achievable without transmitting a single pixel from the target domain.
PLADA establishes a new Pareto frontier for bandwidth efficiency. As illustrated in Figure~\ref{fig:prior_baselines_cub200}, our method (indicated by the star) maintains high accuracy in the extreme low-bandwidth regime ($<1$ MB). Conversely, traditional image-based methods (Random Subset, Coresets) suffer catastrophic accuracy drops in this regime, as they can only transmit a negligible number of training samples.

\paragraph{The ``denoising'' effect of filtering.}
A key finding is that training on a filtered subset (top 1\%--10\% lowest energy) often yields \emph{higher} accuracy than training on the full reference dataset (100\%). For instance, on \emph{FGVC-Aircraft} and \emph{RESISC45}, the filtered subsets significantly outperform the full dataset. This indicates that Energy-based pruning acts as a semantic denoiser: it effectively removes ``distractor'' images that the teacher classifies with low confidence, leaving only the samples that structurally align with the target concepts.

\paragraph{Impact of reference set scale.}
Comparing the two reference sets in Table~\ref{tab:main_results}, the larger ImageNet-21K (14.2M images) consistently yields better downstream performance than ImageNet-1K (1.2M images). The massive diversity of the larger pool increases the probability of finding semantic neighbors for fine-grained target classes, providing a richer training signal.

\begin{table}[ht]
\caption{
\textbf{Results on medical datasets$^{\dagger}$.} ImageNet-21K ref. set.
}
\label{tab:medical_results}
\begin{center}
\begin{small}
\begin{sc}
\setlength{\tabcolsep}{2pt} 
\begin{tabular}{lcccc}
\toprule
Dataset & \begin{tabular}{@{}c@{}}1\% \\ Vanil.\end{tabular} & \begin{tabular}{@{}c@{}}1\% \\ Oppos.\end{tabular} & \begin{tabular}{@{}c@{}}1\%+Safe \\ ($\alpha=0.5$)\end{tabular} & \begin{tabular}{@{}c@{}}1\%+Safe \\ ($\alpha=-0.2$)\end{tabular} \\
\midrule
BloodMNIST       & 18.24\% & \textbf{59.28\%} & 47.00\% & 41.45\% \\
DermaMNIST       & 53.32\% & \textbf{67.68\%} & 47.58\% & 38.05\% \\
RetinaMNIST      & 56.50\% & \textbf{56.75\%} & 55.25\% & 55.00\% \\
NCT-CRC-HE  & 18.69\% & \textbf{43.51\%} & 32.57\% & 32.37\% \\
\bottomrule
\end{tabular}
\end{sc}
\begin{flushleft}
\scriptsize
$^{\dagger}$ \textbf{Teacher Accuracies:} BloodMNIST (99.09\%), DermaMNIST (89.63\%), RetinaMNIST (70.00\%), NCT-CRC-HE-100K (99.93\%).
\end{flushleft}
\end{small}
\end{center}
\vskip -0.1in
\end{table}

\subsection{Analysis}
\label{sec:experiments:ablation_analysis}

\paragraph{The energy paradox in far-OOD tasks (Medical).}
A major challenge arises when the target domain is semantically disjoint from the reference domain. As shown in Table~\ref{tab:medical_results}, standard low-energy filtering fails for medical datasets. In these cases, the ``best'' reference images (lowest energy) are often generic natural images that map spuriously to a single target class (e.g., a red circle in ImageNet mapped to a blood cell), causing the student model to collapse. However, we observe a reversal in the optimal strategy: selecting images with the \textit{highest energy} (highest uncertainty) consistently outperforms standard filtering (Table~\ref{tab:medical_results}, \textit{Inverse} column).
We hypothesize that high-energy reference images, likely containing high-frequency patterns or unusual textures, possess low-level structural statistics that align better with medical scans than semantically clear natural images. This suggests an adaptive strategy: utilize low-energy selection for in-domain tasks and high-energy (inverse) selection for far-OOD tasks.

\paragraph{Safety-Net Filtering.} Standard energy filtering can disproportionately prune hard-to-classify or under-represented categories. Table~\ref{tab:safetynet_results} demonstrates the efficacy of our Safety-Net mechanism ($\alpha=-0.2$), which enforces a quota for tail classes. For datasets with high inter-class imbalance, such as \emph{RESISC45}, Safety-Net filtering significantly boosts accuracy (from 58.16\% to 75.65\% at 1\% keep-rate) by ensuring the student receives a balanced training distribution even under extreme compression.

\paragraph{Payload Compression Analysis.}
We analyze the impact of variable-length coding on the final payload size in Table~\ref{tab:compression_results}.
\begin{enumerate}
    \item \textit{Sparsity exploitation:} At strict filtering rates ($p \le 1\%$), the payload is dominated by the indices of the selected images rather than the labels themselves.
    \item \textit{Compression strategy:} Zstandard (Zstd) outperforms Huffman coding by exploiting local correlations in the sparse index sequences.
\end{enumerate}
This optimization reduces the total payload for the 1\% setting to between \textit{45 KB and 200 KB}, confirming that transmitting a massive 14-million-image training signal is feasible over even the most constrained channels (e.g., deep-sea acoustic links).

\section{Discussion}

\paragraph{Runtime.} When using high keep ratios (e.g., $p \geq 25\%$), training the student model can take up to 3 days on a single A5000 GPU. However, at low keep ratios the experiments are much shorter (e.g., $\sim$20 minutes at $p=1\%$ with ImageNet-21K as the reference set).

\begin{figure}[ht]
  \vskip 0.2in
  \begin{center}
    \centerline{\includegraphics[width=\columnwidth]{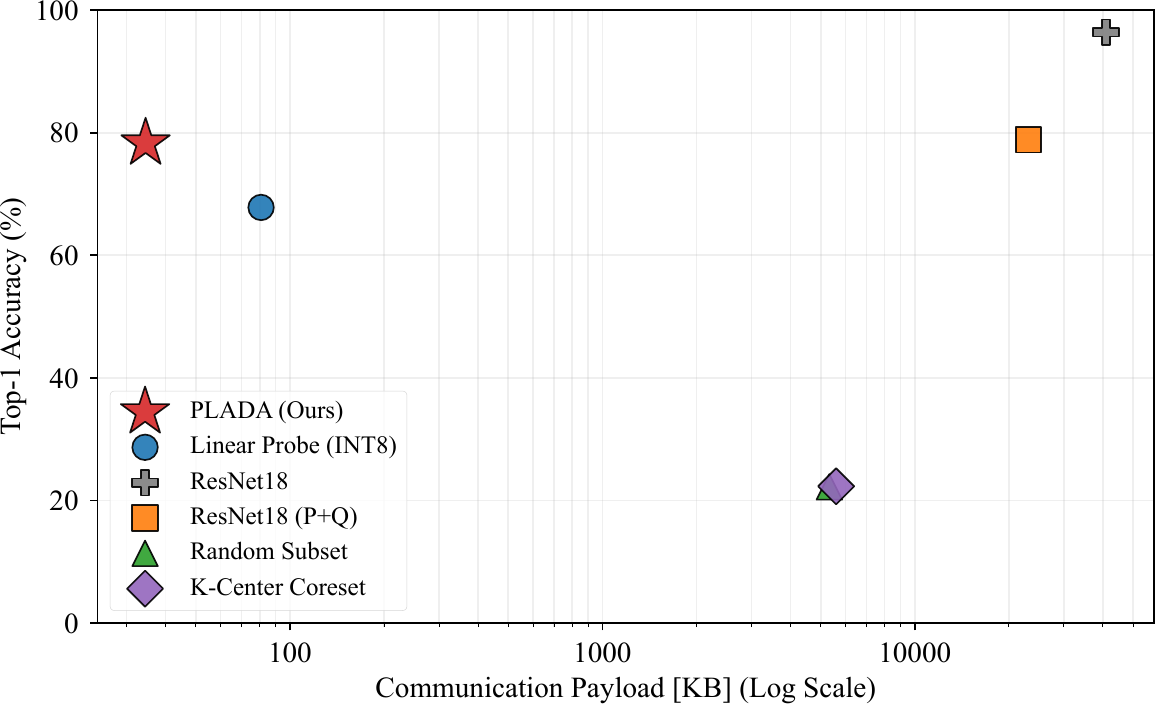}}
    \caption{
        \textbf{Bandwidth-Accuracy Baselines (CUB-200).} Comparison of PLADA against weight and data transmission baselines. PLADA (red star) dominates the top-left corner, achieving \emph{higher} accuracy than weight-based methods while requiring a smaller payload (<35 KB). Data-centric baselines (Random Subset/K-Center) fail to provide a viable signal at this extreme budget. All payloads are Zstd-compressed (level 19).
    }
    \label{fig:prior_baselines_cub200}
  \end{center}
\end{figure}

\paragraph{Transmitting model weights.} While we focus on transmitting datasets as they allow each client to train their model of choice, there are cases where we may consider sending the weights of a given model to clients. We tested several such strategies: (i) training a linear probing model based on a frozen backbone encoder, then sending its INT8 encoded weights. (ii) Sending a ResNet-18 teacher model with optional pruning and INT8 quantization (following model-compression ideas such as Deep Compression~\cite{han2015DeepCompressionPruningQuantization}). We present results on CUB-200 in ~\cref{fig:prior_baselines_cub200}. We observe that the linear probe is the most efficient baseline and is quite accurate. Sending the full model is far more expensive than our approach for sending labels.

\paragraph{Optimal reference dataset selection.} In our experiments, we used ImageNet-1K and ImageNet-21K as reference datasets. These are not necessarily optimal from an accuracy--bandwidth--storage perspective. We are not aware of principled approaches for selecting an optimal reference dataset, and we leave this to future work.

\paragraph{Limitations.} While significantly reducing communication cost, our method requires each client to store the reference dataset. This overhead becomes less significant---and can even be cost-saving---once many target tasks are served and their cumulative size exceeds that of the reference dataset. Another limitation is that, in some cases, training time may increase (i.e., more iterations may be needed) to match training on the original target data. Finally, our work focuses on classification and does not yet handle regression or generative tasks. We expect regression to be straightforward to incorporate, but enabling generative modeling without sending pixels remains an exciting challenge for the future.

\section{Conclusion}
\label{sec:conclusion}
We proposed Pseudo-Labels as Data (PLADA), a method for sending datasets at very low communication cost. It transmits tasks by only sending the hard pseudo-labels for a large, preloaded reference dataset. By combining energy-based filtering with a safety-net mechanism, PLADA selects a compact, class-preserving subset of reference images while aggressively reducing transmission cost. This enables task transfer with payloads well below 1 MB, even when using huge ImageNet-21K reference set. These results show that, for classification, task knowledge can be conveyed more efficiently through labels rather than through pixels. We hope this perspective motivates future work to further improve the accuracy--bandwidth trade-off in dataset serving.


\nocite{paszke2019pytorch,wightman2019timm}

\bibliography{main}
\bibliographystyle{icml2026}

\newpage
\appendix
\onecolumn

\section{Dataset Intersection Analysis}
\label{sec:app:intersection}

To validate our method, we must ensure that the student model is not simply memorizing samples from the reference dataset (ImageNet-21K) that happen to be duplicates of the target task's test set. We implemented a content-based intersection check using the method described below.

\textbf{Methodology:} We employed a bucketed L1-distance check. We computed the mean and variance for every image in both the target datasets (Set 1) and the reference ImageNet-21K dataset (Set 2). Images were hashed into buckets based on these statistics (using 1024 bins). We then performed a pixel-wise L1 comparison only between images falling into the same buckets. An intersection was flagged if the mean L1 pixel difference was below a strict threshold ($\epsilon < 10^{-5}$).

\textbf{Results:} We analyzed the intersection between the target datasets and the full 14.2M ImageNet-21K dataset. Our findings are summarized in Table \ref{tab:intersections}.

\begin{itemize}
    \item \textbf{Zero Intersection:} For the majority of benchmarks—including Oxford Flowers 102, Food-101, DTD, CIFAR-10, RESISC45, and Places365—we found exactly \textbf{0} intersections between the target and test sets with ImageNet-21K.
    \item \textbf{Negligible Intersection:}
    \begin{itemize}
        \item \textbf{FGVC-Aircraft \& CUB-200-2011:} While we identified a small number of duplicates in the training splits (1 and 2 images, respectively), the intersection with the \textit{test sets} was exactly \textbf{0}.
        \item \textbf{Caltech-101:} We found 2 overlapping images in the test set.
        \item \textbf{Oxford-IIIT Pet:} This dataset showed the highest overlap, with 25 test images appearing in ImageNet-21K. However, this represents only $\approx 0.68\%$ of the test set (25/3669).
    \end{itemize}
\end{itemize}

Given that the intersections are either non-existent or statistically negligible ($<1\%$), we conclude that the performance gains reported in our experiments are driven by the effective distillation of knowledge into the synthetic labels, rather than data leakage from the auxiliary set.

\begin{table}[h]
\caption{Intersection analysis between Target Datasets and ImageNet-21K (14.2M). Columns show the number of duplicates found in the full target dataset ($x/n_{all}$) and the specific test set ($y/n_{test}$).}
\label{tab:intersections}
\begin{center}
\begin{small}
\begin{sc}
\begin{tabular}{lcc}
\toprule
Target Dataset & Total Intersection & Test Intersection \\
\midrule
Oxford Flowers 102 & 0 / 8,189 & 0 / 6,149 \\
Food-101           & 0 / 101,000 & 0 / 25,250 \\
DTD                & 0 / 3,760 & 0 / 1,880 \\
CIFAR-10           & 0 / 60,000 & 0 / 10,000 \\
RESISC45           & 0 / 31,500 & 0 / 6,300 \\
Places365          & 0 / 1,839,960 & 0 / 36,500 \\
\midrule
FGVC-Aircraft      & 1 / 10,000 & 0 / 3,333 \\
CUB-200-2011       & 2 / 11,788 & 0 / 2,358 \\
Caltech-101        & 4 / 8,677 & 2 / 1,736 \\
Oxford-IIIT Pet    & 244 / 7,349 & 25 / 3,669 \\
\bottomrule
\end{tabular}
\end{sc}
\end{small}
\end{center}
\end{table}

\section{Detailed Experimental Results}
\label{sec:app:experiment}
We report detailed student accuracy results under different filtering strategies in \cref{tab:entropy_vs_energy,tab:opposite_energy,tab:opposite_energy_med,tab:safetynet_imagenet_21k_vs_1k,tab:medical_safetynet_imagenet_21k_vs_1k}. ~\cref{tab:entropy_vs_energy} compares entropy-based and energy-based filtering across multiple filtering budgets using ImageNet-21K as the reference dataset. ~\cref{tab:opposite_energy} evaluates an alternative setting in which the student is trained using the highest $p\%$ energy-score images (instead of the lowest), for models trained with ImageNet-1K or ImageNet-21K as reference datasets. ~\cref{tab:opposite_energy_med} presents the same analysis for biomedical target datasets, highlighting the behavior of highest-energy versus lowest-energy filtering in this out-of-distribution setting. \cref{tab:safetynet_imagenet_21k_vs_1k,tab:medical_safetynet_imagenet_21k_vs_1k} report Safety-Net results on natural image and medical datasets, respectively.

\begin{table}[H]
\centering
\caption{
\textbf{Entropy vs Energy Pruning.} This table reports student accuracy$^{\dagger}$ under different filtering budgets, comparing entropy-based pruning with energy-based pruning (ImageNet-21K).
}
\label{tab:entropy_vs_energy}
\vskip 0.15in
\begin{small}
\setlength{\tabcolsep}{1.4pt}
\begin{sc}
\begin{tabular}{l @{\hspace{6pt}} cc @{\hspace{8pt}} cc @{\hspace{8pt}} cc @{\hspace{8pt}} cc}
\toprule
\abovespace
& \multicolumn{2}{c}{1\% Filter} & \multicolumn{2}{c}{5\% Filter} & \multicolumn{2}{c}{10\% Filter} & \multicolumn{2}{c}{25\% Filter} \\
\cmidrule(lr){2-3} \cmidrule(lr){4-5} \cmidrule(lr){6-7} \cmidrule(lr){8-9}
Dataset & Entropy & Energy & Entropy & Energy & Entropy & Energy & Entropy & Energy \\
\midrule
\abovespace
Caltech-101 & 75.06\% & \textbf{79.84\%} & 84.22\% & \textbf{88.94\%} & 87.56\% & \textbf{90.21\%} & \textbf{91.71\%} & 90.73\% \\
CIFAR-10 & \textbf{72.57\%} & 63.31\% & \textbf{86.00\%} & 85.31\% & \textbf{89.14\%} & 88.12\% & 91.14\% & \textbf{91.68\%} \\
CUB-200 & 77.44\% & \textbf{82.49\%} & 81.42\% & \textbf{82.36\%} & 81.76\% & \textbf{82.44\%} & 81.59\% & \textbf{81.34\%} \\
DTD & \textbf{67.02\%} & 66.65\% & 69.79\% & \textbf{70.16\%} & \textbf{71.22\%} & 70.69\% & \textbf{71.06\%} & 70.80\% \\
FGVC-Aircraft & 37.32\% & \textbf{53.62\%} & 44.13\% & \textbf{45.51\%} & 45.69\% & \textbf{46.41\%} & \textbf{45.90\%} & 45.87\% \\
Food-101 & 68.32\% & \textbf{75.50\%} & 73.88\% & \textbf{76.18\%} & 74.33\% & \textbf{76.91\%} & 75.47\% & \textbf{76.72\%} \\
Oxford-Flowers-102 & 95.01\% & \textbf{96.93\%} & 98.28\% & \textbf{98.41\%} & 98.41\% & \textbf{98.50\%} & 98.18\% & \textbf{98.19\%} \\
Oxford-IIIT-Pet & 90.81\% & \textbf{90.95\%} & \textbf{91.09\%} & 91.03\% & 90.76\% & \textbf{90.81\%} & \textbf{90.41\%} & 90.05\% \\
Places365 & 20.38\% & \textbf{23.39\%} & 34.71\% & \textbf{34.89\%} & \textbf{40.40\%} & 40.05\% & 46.20\% & 45.82\% \\
RESISC45 & 54.05\% & \textbf{58.16\%} & \textbf{70.11\%} & 67.81\% & \textbf{75.76\%} & 74.37\% & \textbf{80.87\%} & 80.62\% \\
\bottomrule
\end{tabular}
\end{sc}
\begin{flushleft}
\scriptsize
$^{\dagger}$ \textbf{Teacher Accuracies:} Caltech-101 (98.39\%), CIFAR-10 (98.15\%), CUB-200 (97.71\%), DTD (77.50\%), FGVC-Aircraft (86.53\%), Food-101 (90.02\%), Oxford-Flowers (99.04\%), Oxford-Pets (93.40\%), Places365 (55.45\%), RESISC45 (96.84\%).
\end{flushleft}
\end{small}
\end{table}

\begin{table}[H]
\centering
\caption{
\textbf{Opposite Energy-Based Pruning.} This table shows the student accuracy obtained by training on the \textbf{highest}-$p\%$ energy-score images (which are usually the "worst" images), using ImageNet-1K and ImageNet-21K as reference datasets.
}
\label{tab:opposite_energy}
\vskip 0.15in
\begin{small}
\setlength{\tabcolsep}{5pt}
\begin{sc}
\begin{tabular}{l @{\hspace{8pt}} ccc @{\hspace{8pt}} cccc}
\toprule
\abovespace
& \multicolumn{3}{c}{ImageNet-21K (14.2M images)} & \multicolumn{4}{c}{ImageNet-1K (1.2M images)} \\
\cmidrule(lr){2-4} \cmidrule(lr){5-8}
Dataset & 1\% & 5\% & 10\% & 1\% & 5\% & 10\% & 25\% \\
\midrule
\abovespace
Caltech-101 & 74.42\% & 85.66\% & \textbf{87.27\%} & 61.29\% & 79.32\% & 82.89\% & \textbf{85.54\%} \\
CIFAR-10 & 54.80\% & 64.45\% & \textbf{70.12\%} & 34.90\% & 45.84\% & 58.46\% & \textbf{67.09\%} \\
CUB-200 & 6.19\% & 11.79\% & \textbf{15.48\%} & 3.22\% & 6.66\% & 8.78\% & \textbf{14.16\%} \\
DTD & 26.12\% & 37.34\% & \textbf{42.77\%} & 19.41\% & 27.45\% & 35.90\% & \textbf{40.00\%} \\
FGVC-Aircraft & 3.45\% & 5.37\% & \textbf{8.16\%} & 2.46\% & 3.57\% & 4.74\% & \textbf{8.70\%} \\
Food-101 & 6.28\% & 13.11\% & \textbf{20.11\%} & 3.72\% & 7.40\% & 11.30\% & \textbf{18.15\%} \\
Oxford-Flowers-102 & 9.95\% & 24.43\% & \textbf{29.91\%} & 4.36\% & 10.88\% & 14.41\% & \textbf{24.15\%} \\
Oxford-IIIT-Pet & 18.67\% & 30.23\% & \textbf{36.49\%} & 9.59\% & 19.81\% & 29.05\% & \textbf{37.20\%} \\
Places365 & 18.04\% & 27.35\% & \textbf{32.14\%} & 10.91\% & 18.96\% & 22.73\% & \textbf{29.55\%} \\
RESISC45 & 2.06\% & 2.06\% & 2.06\% & 29.65\% & 52.43\% & 61.33\% & \textbf{68.62\%} \\
\bottomrule
\end{tabular}
\end{sc}
\end{small}
\end{table}

\begin{table}[H]
\centering
\caption{
\textbf{Medical Datasets Opposite Energy-Based Pruning.}
This table presents student accuracy obtained by training on the highest $p\%$ energy-score images of the reference sets. As explained in \cref{sec:experiments:ablation_analysis}, we unexpectedly observed higher student accuracy when applying the opposite filtering strategy on the medical datasets.
}
\label{tab:opposite_energy_med}
\vskip 0.15in
\begin{small}
\setlength{\tabcolsep}{5pt}
\begin{sc}
\begin{tabular}{l @{\hspace{8pt}} cc @{\hspace{8pt}} cc}
\toprule
\abovespace
& \multicolumn{2}{c}{ImageNet-21K (14.2M images)} & \multicolumn{2}{c}{ImageNet-1K (1.2M images)} \\
\cmidrule(lr){2-3} \cmidrule(lr){4-5}
Dataset & 1\% & 5\% & 1\% & 5\% \\
\midrule
\abovespace
BloodMNIST & \textbf{59.28\%} & 57.88\% & 32.65\% & \textbf{38.03\%} \\
DermaMNIST & \textbf{67.68\%} & 66.58\% & 66.43\% & \textbf{67.33\%} \\
RetinaMNIST & 56.75\% & \textbf{58.00\%} & 50.25\% & \textbf{55.50\%} \\
NCT-CRC-HE-100K & 43.51\% & \textbf{52.48\%} & 26.46\% & \textbf{35.33\%} \\
\bottomrule
\end{tabular}
\end{sc}
\end{small}
\end{table}

Notably, the results for the medical datasets differ from those of the other benchmarks. This behavior can be attributed to the significant domain gap between these datasets and both the natural-image benchmarks and ImageNet. As a consequence, our filtering strategy has a limited effect in this setting.

\begin{table}[H]
\centering
\caption{
\textbf{Safety-net Filtering.} This table shows student accuracy when using Safety-Net filtering with $\alpha=-0.2, 0.5$ across different filtering keep ratios, with ImageNet-21K and ImageNet-1K as reference sets.
}
\label{tab:safetynet_imagenet_21k_vs_1k}
\vskip 0.15in
\setlength{\tabcolsep}{2pt}
\begin{sc}
{\fontsize{8}{9.5}\selectfont
\begin{tabular*}{\textwidth}{@{\extracolsep{\fill}} l c c c c c c c c c c c c}
\toprule
\abovespace
& \multicolumn{4}{c}{ImageNet-21K} & \multicolumn{8}{c}{ImageNet-1K} \\
\cmidrule(lr){2-5} \cmidrule(lr){6-13}

& \multicolumn{2}{c}{1\%} & \multicolumn{2}{c}{5\%}
& \multicolumn{2}{c}{1\%} & \multicolumn{2}{c}{5\%} & \multicolumn{2}{c}{10\%} & \multicolumn{2}{c}{25\%} \\
\cmidrule(lr){2-3} \cmidrule(lr){4-5}
\cmidrule(lr){6-7} \cmidrule(lr){8-9} \cmidrule(lr){10-11} \cmidrule(lr){12-13}

Dataset
& $\alpha\!=\!-0.2$ & $\alpha\!=\!0.5$
& $\alpha\!=\!-0.2$ & $\alpha\!=\!0.5$
& $\alpha\!=\!-0.2$ & $\alpha\!=\!0.5$
& $\alpha\!=\!-0.2$ & $\alpha\!=\!0.5$
& $\alpha\!=\!-0.2$ & $\alpha\!=\!0.5$
& $\alpha\!=\!-0.2$ & $\alpha\!=\!0.5$ \\
\midrule
\abovespace
Caltech-101 & 86.29\% & \textbf{86.69\%} & 90.67\% & \textbf{91.07\%} & \textbf{77.25\%} & \textbf{77.25\%} & 83.41\% & \textbf{83.81\%} & \textbf{86.81\%} & 85.20\% & \textbf{86.06\%} & 85.83\% \\
CIFAR-10 & 74.62\% & \textbf{76.75\%} & 83.83\% & \textbf{85.44\%} & 58.07\% & \textbf{58.59\%} & \textbf{69.97\%} & 68.27\% & \textbf{73.51\%} & 72.56\% & 77.00\% & \textbf{78.26\%} \\
CUB-200 & 80.53\% & \textbf{81.21\%} & \textbf{82.06\%} & 82.06\% & 15.48\% & \textbf{15.78\%} & \textbf{41.26\%} & 39.61\% & \textbf{52.54\%} & 52.50\% & 48.39\% & \textbf{49.15\%} \\
DTD & \textbf{68.09\%} & 66.70\% & \textbf{70.59\%} & 70.48\% & \textbf{53.62\%} & 51.65\% & \textbf{60.16\%} & 60.11\% & 61.81\% & \textbf{62.29\%} & 61.28\% & \textbf{62.77\%} \\
FGVC-Aircraft & \textbf{44.58\%} & 43.23\% & \textbf{46.56\%} & 45.66\% & 10.29\% & \textbf{10.35\%} & 18.15\% & \textbf{18.24\%} & \textbf{27.00\%} & 26.82\% & 21.39\% & \textbf{22.50\%} \\
Food-101 & \textbf{71.66\%} & 70.91\% & \textbf{76.70\%} & 75.94\% & \textbf{24.59\%} & 22.51\% & \textbf{43.52\%} & 42.07\% & \textbf{54.23\%} & 53.81\% & \textbf{48.72\%} & 48.41\% \\
Oxford-Flowers-102 & 97.35\% & \textbf{97.53\%} & 98.42\% & \textbf{98.44\%} & \textbf{38.61\%} & 33.44\% & 62.21\% & \textbf{64.25\%} & 73.83\% & \textbf{74.70\%} & 64.53\% & \textbf{76.00\%} \\
Oxford-IIIT-Pet & 90.87\% & \textbf{90.98\%} & 90.87\% & \textbf{91.11\%} & \textbf{84.66\%} & 84.08\% & \textbf{89.21\%} & 88.91\% & \textbf{89.64\%} & 89.23\% & 89.45\% & \textbf{89.53\%} \\
Places365 & \textbf{31.59\%} & 30.26\% & \textbf{39.21\%} & 38.33\% & \textbf{18.00\%} & 15.84\% & \textbf{30.14\%} & 29.23\% & \textbf{35.71\%} & 35.14\% & \textbf{37.90\%} & 37.32\% \\
RESISC45 & \textbf{75.65\%} & 72.81\% & \textbf{82.06\%} & 79.83\% & \textbf{44.84\%} & 36.32\% & \textbf{68.62\%} & 63.95\% & \textbf{76.52\%} & 73.44\% & \textbf{75.56\%} & 74.02\% \\
\bottomrule
\end{tabular*}
}
\end{sc}
\end{table}

\begin{table}[H]
\centering
\caption{
\textbf{Medical Datasets Safety-net Filtering.} This table shows student accuracy when using Safety-Net filtering with $\alpha=-0.2, 0.5$ across different filtering keep ratios, with ImageNet-21K and ImageNet-1K as reference sets.
}
\label{tab:medical_safetynet_imagenet_21k_vs_1k}
\vskip 0.15in
\setlength{\tabcolsep}{2pt}
\begin{sc}
\begin{tabular*}{\textwidth}{@{\extracolsep{\fill}} l c c c c c c c c}
\toprule
\abovespace
& \multicolumn{4}{c}{ImageNet-21K} & \multicolumn{4}{c}{ImageNet-1K} \\
\cmidrule(lr){2-5} \cmidrule(lr){6-9}

& \multicolumn{2}{c}{1\%} & \multicolumn{2}{c}{5\%}
& \multicolumn{2}{c}{1\%} & \multicolumn{2}{c}{5\%}  \\
\cmidrule(lr){2-3} \cmidrule(lr){4-5}
\cmidrule(lr){6-7} \cmidrule(lr){8-9}

Dataset
& $\alpha\!=\!-0.2$ & $\alpha\!=\!0.5$
& $\alpha\!=\!-0.2$ & $\alpha\!=\!0.5$
& $\alpha\!=\!-0.2$ & $\alpha\!=\!0.5$
& $\alpha\!=\!-0.2$ & $\alpha\!=\!0.5$ \\
\midrule
\abovespace
BloodMNIST & 41.45\% & \textbf{47.00\%} & \textbf{56.71\%} & 45.19\% & 38.91\% & \textbf{42.03\%} & 45.31\% & \textbf{48.44\%} \\
DermaMNIST & 38.05\% & \textbf{47.58\%} & 68.33\% & \textbf{68.33\%} & 51.67\% & \textbf{53.42\%} & 57.06\% & \textbf{57.91\%} \\
NCT-CRC-HE-100K & 32.37\% & \textbf{32.57\%} & \textbf{41.89\%} & 37.54\% & 23.79\% & \textbf{28.23\%} & 33.88\% & \textbf{39.47\%} \\
RetinaMNIST & 55.00\% & \textbf{55.25\%} & \textbf{63.25\%} & 61.25\% & 53.75\% & \textbf{55.75\%} & \textbf{58.50\%} & 58.25\% \\
\bottomrule
\end{tabular*}
\end{sc}
\end{table}

\section{Energy Filtering Visualizations}
\label{sec:app:energy_visualizations}
Energy-based filtering is our primary mechanism for selecting a small, informative subset of the reference set.
For each reference image $x$, we score it using the teacher trained on the target dataset and compute its logit energy (\cref{eq:energy_formula}), where \emph{lower} energy indicates a more confident and concentrated prediction over the target classes.
We then rank all reference images by energy and keep only the lowest $p\%$.
Intuitively, this procedure removes reference images for which the teacher produces diffuse (high-uncertainty) logits, which are unlikely to correspond to any target concept and would otherwise introduce noisy pseudo-labels.

\begin{figure*}[p]
  \vskip 0.2in
  \begin{center}
    \includegraphics[width=\textwidth]{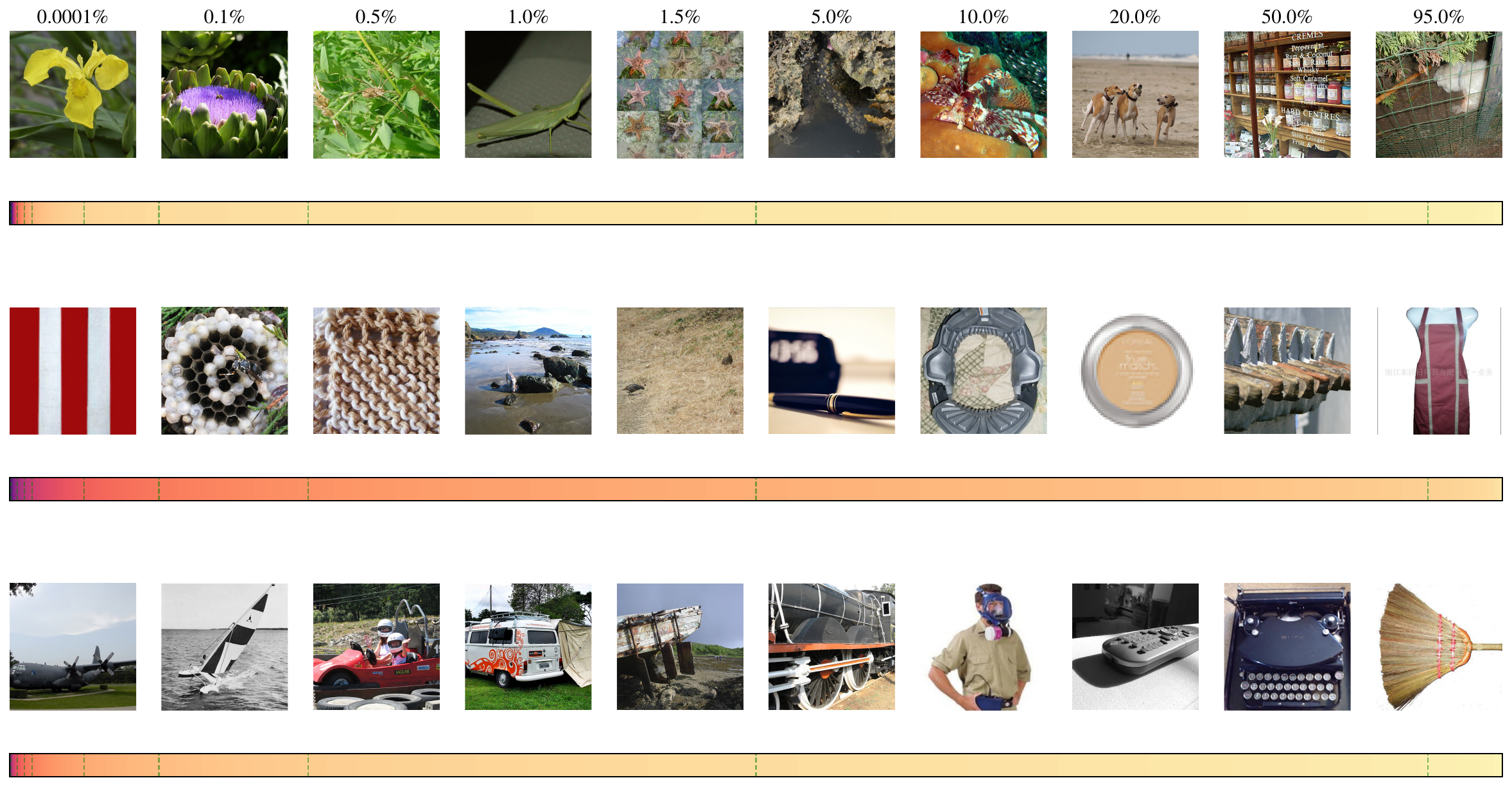}
    \caption{
        \textbf{Energy percentiles on ImageNet-1K.}
        For three target tasks (rows: Oxford-Flowers102, DTD, FGVC-Aircraft), we score every ImageNet-1K image using the corresponding target teacher and sort the reference set by logit energy (\cref{eq:energy_formula}; lower is better).
        We then show exemplar reference images at fixed energy percentiles (columns).
        The horizontal bar under each row visualizes the full energy range over the entire reference set (dashed ticks mark the sampled percentiles).
        As the percentile increases (left$\rightarrow$right), samples transition from target-aligned content (flowers/texture patterns/aircraft) to increasingly irrelevant images.
    }
    \label{fig:filtering_analysis__ILSVRC2012_oxford_flowers102_dtd_fgvc_aircraft}
  \end{center}
\end{figure*}

\begin{figure*}[p]
  \vskip 0.2in
  \begin{center}
    \includegraphics[width=\textwidth]{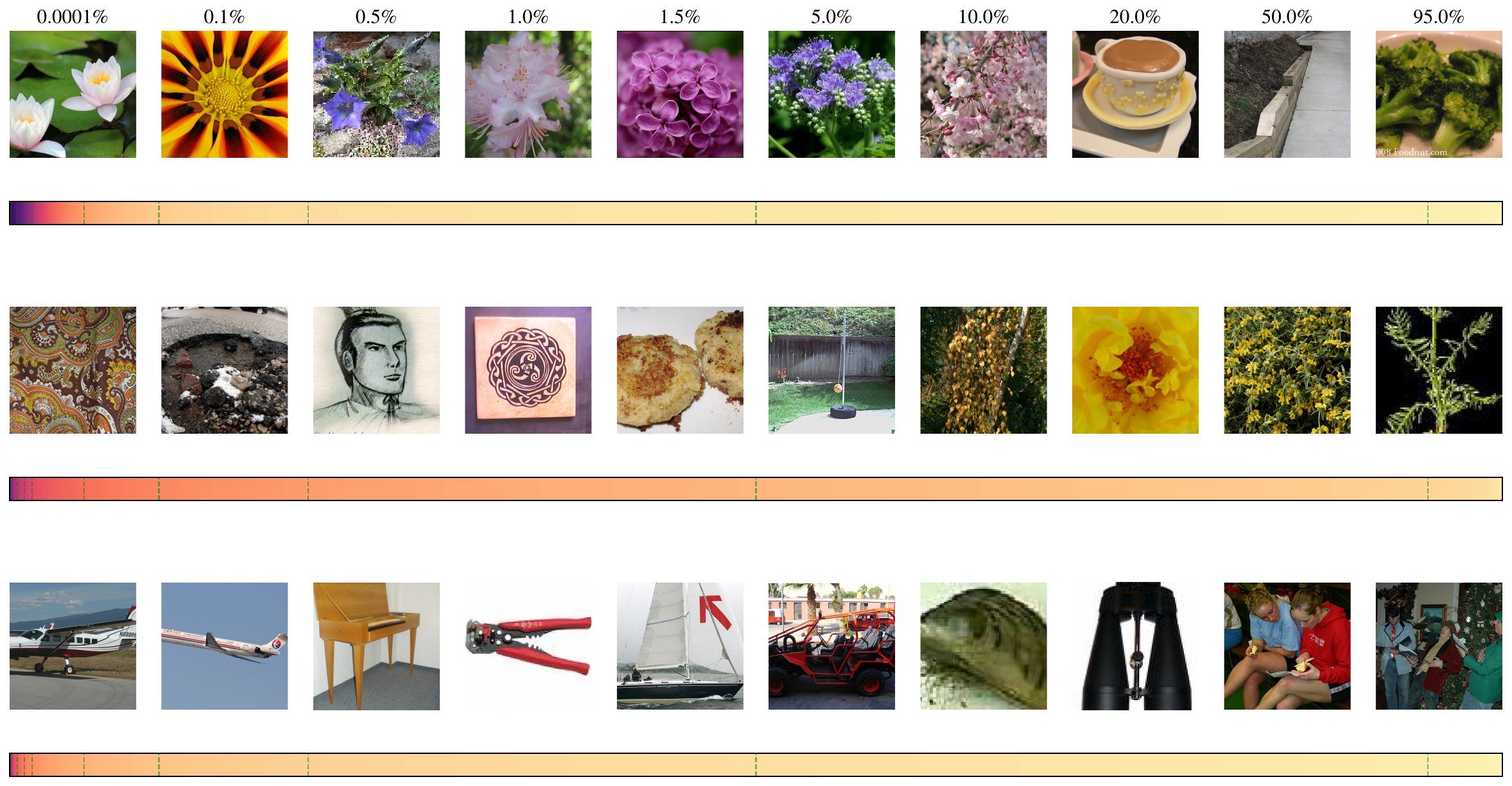}
    \caption{
        \textbf{Energy percentiles on ImageNet-21K.} Same visualization as \cref{fig:filtering_analysis__ILSVRC2012_oxford_flowers102_dtd_fgvc_aircraft} but using ImageNet-21K (14.2M images) as the reference set. The larger and more diverse dataset typically provides closer semantic neighbors in the low-energy tail (e.g., more flower varieties and texture-like patterns).
    }
    \label{fig:filtering_analysis__ImageNet_21K__oxford_flowers102_dtd_fgvc_aircraft}
  \end{center}
\end{figure*}

\cref{fig:filtering_analysis__ILSVRC2012_oxford_flowers102_dtd_fgvc_aircraft,fig:filtering_analysis__ImageNet_21K__oxford_flowers102_dtd_fgvc_aircraft} visualize this ranking for three representative targets (Oxford-Flowers102, DTD, and FGVC-Aircraft).
Each row fixes the target teacher, and columns sweep over energy percentiles (left$\rightarrow$right).
The low-energy tail (left) is dominated by images that are visually and semantically aligned with the target domain---e.g., flower close-ups for Oxford-Flowers102, texture-like patterns for DTD, and aircraft/nearby vehicle imagery for FGVC-Aircraft.
As we move toward higher percentiles (right), the samples become increasingly unrelated, illustrating that retaining high-energy images would primarily add label noise.
Comparing the two reference sets, ImageNet-21K typically yields closer semantic neighbors than ImageNet-1K, reflecting its larger scale and diversity.

\begin{figure*}[p]
  \vskip 0.2in
  \begin{center}
    \includegraphics[width=\textwidth]{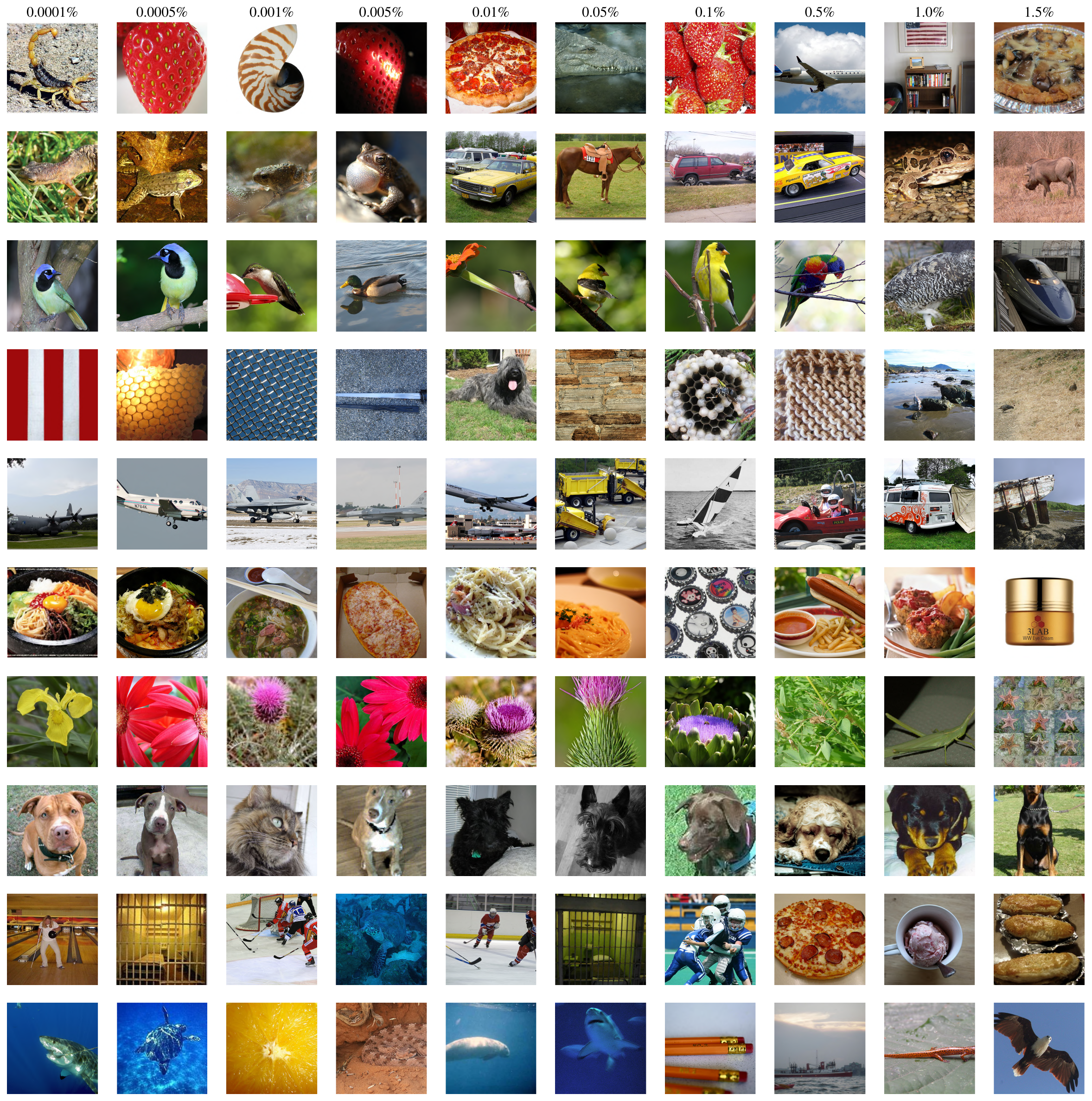}
    \caption{
        \textbf{Low-energy reference images (ImageNet-1K).}
        For each target teacher (rows; top-to-bottom: Caltech-101, CIFAR-10, CUB-200, DTD, FGVC-Aircraft, Food-101, Oxford-Flowers, Oxford-Pets, Places365, RESISC45), we show ImageNet-1K reference images drawn from increasingly larger low-energy percentiles (columns: 0.0001\%--1.5\%).
        The extreme low-energy tail tends to contain canonical instances of the target concepts (e.g., birds for CUB-200, food dishes for Food-101, flowers for Oxford-Flowers).
    }
    \label{fig:filtering_analysis_ILSVRC2012_grid_low}
  \end{center}
\end{figure*}

\begin{figure*}[p]
  \vskip 0.2in
  \begin{center}
    \includegraphics[width=\textwidth]{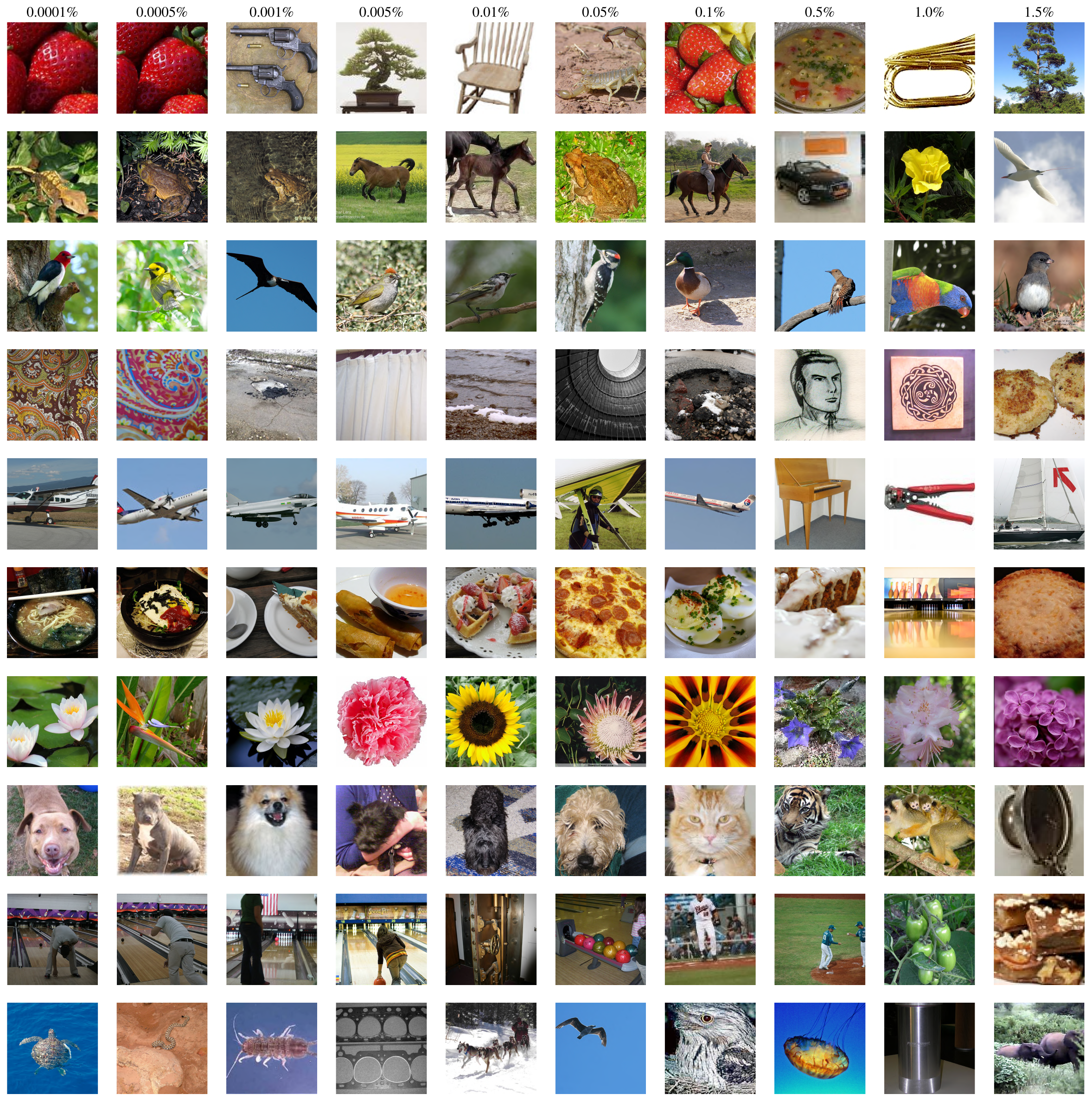}
    \caption{
        \textbf{Low-energy reference images (ImageNet-21K).}
        Same as \cref{fig:filtering_analysis_ILSVRC2012_grid_low} but using ImageNet-21K as the reference set.
        The larger reference set yields a richer and often more semantically aligned set of low-energy exemplars across targets, consistent with the higher student accuracy obtained with ImageNet-21K as the reference set in \cref{tab:main_results}.
    }
    \label{fig:filtering_analysis_ImageNet_21K_grid_low}
  \end{center}
\end{figure*}

\cref{fig:filtering_analysis_ILSVRC2012_grid_low,fig:filtering_analysis_ImageNet_21K_grid_low} zoom into the extreme low-energy region for all ten natural-image benchmarks.
Across datasets, the retrieved images at very small percentiles look like canonical exemplars of the target concepts (birds for CUB-200, food dishes for Food-101, flowers for Oxford-Flowers, pets for Oxford-Pets, etc.).
This qualitative behavior helps explain why aggressive keep rates (e.g., $1\%$ or below) can still provide a strong training signal: the filter concentrates the transmitted supervision on the subset of reference images that the teacher regards as most in-distribution for the target task.

\clearpage 

\section{Extended Methodology}
\label{sec:app:methodology}
In this appendix, we provide mathematical formulations and implementations for the additional filtering, labeling, and training strategies investigated in this work. Although these methods demonstrated reasonable performance, they did not outperform the primary PLADA method introduced in the main text.

\subsection{Uncertainty Metrics}
Let $f_\theta(\mathbf{x}) \in \mathbb{R}^C$ denote the logits output by the teacher model for an input $\mathbf{x}$, and let $T$ be the temperature scaling parameter.

\paragraph{Energy Score.} 
We utilize the free energy function, commonly used for out-of-distribution detection \cite{liu2020energy}. The energy maps the logit distribution to a scalar value, where lower energy implies higher likelihood (higher confidence):
\begin{equation}
    E(\mathbf{x}; T) = -T \cdot \log \sum_{j=1}^C \exp\left(\frac{f_\theta(\mathbf{x})_j}{T}\right)
\end{equation}

\paragraph{Entropy Score.} 
We compute the Shannon entropy of the predictive distribution obtained via the softmax function, $\mathbf{p} = \sigma(f_\theta(\mathbf{x})/T)$:
\begin{equation}
    H(\mathbf{x}) = -\sum_{j=1}^C p_j \log p_j
\end{equation}
where higher entropy indicates higher uncertainty (lower confidence).

\subsection{Filtering Strategies}

\subsubsection{Rank Normalization}
Directly comparing raw scores (e.g., Energy vs. Entropy) is difficult due to differing scales and distributions. We therefore convert scores into normalized ranks. Let $\mathcal{S} = \{s_1, \dots, s_N\}$ be the raw scores for the entire dataset. The normalized rank $r_i \in [0, 1]$ for image $i$ is defined as:
\begin{equation}
    r_i = \frac{\text{rank}(s_i)}{N - 1}
\end{equation}
where $\text{rank}(s_i)$ is the 0-based index of $s_i$ in the sorted array of scores (ascending order, such that $r_i=0$ represents the "best" score, i.e., lowest uncertainty).

\subsubsection{Consensus (Intersection) Filtering}
To combine multiple filtering criteria (denoted $M_1, \dots, M_m$), we seek samples that are highly ranked across \textit{all} methods. We define the consensus cost $c_i$ for image $i$ as the maximum normalized rank assigned by any constituent method:
\begin{equation}
    c_i = \max_{k=1}^m \left( r_i^{(M_k)} \right)
\end{equation}
We then select the subset of indices $\mathcal{I}_{keep}$ corresponding to the smallest values $c_i$ such that $|\mathcal{I}_{keep}| = \lfloor N \cdot \beta \rfloor$, where $\beta$ is the keep ratio. This intersection strategy ensures that any selected image belongs to the top percentile of \textit{every} applied filter.

\subsection{Labeling Variants}
Standard Knowledge Distillation (KD) \cite{Hinton2015DistillingTK} is typically more effective when using soft labels \cite{qin2024LabelWorth1000Images}; however, this approach incurs a significant transmission overhead compared to hard labels. We explored two methods to approximate the benefits of soft labels while maintaining the low payload cost associated with hard labels. Ultimately, these methods did not outperform the standard use of hard labels.

\subsubsection{Average Soft-Labels}
To capture inter-class similarities, we compute a global prototype for each hard class $c \in \{1, \dots, C\}$. Let $\mathcal{D}_c$ be the set of all proxy images assigned to hard label $c$. The average soft label $\bar{\mathbf{y}}_c \in \mathbb{R}^C$ is:
\begin{equation}
    \bar{\mathbf{y}}_c = \frac{1}{|\mathcal{D}_c|} \sum_{\mathbf{x} \in \mathcal{D}_c} \sigma(f_\theta(\mathbf{x}))
\end{equation}
During training, if a student image has hard label $c$, it trains against the static target $\bar{\mathbf{y}}_c$.

\subsubsection{Dirichlet Distribution Estimation}
To model intra-class variance without storing per-sample targets, we assume the soft labels for class $c$ follow a Dirichlet distribution, $\mathbf{y} \sim \text{Dir}(\boldsymbol{\alpha}_c)$. We estimate the concentration parameters $\boldsymbol{\alpha}_c$ using the Method of Moments.

For a specific class $c$, let $\mu_j$ and $\sigma_j^2$ be the empirical mean and variance of the probability $p_j$ across all images in $\mathcal{D}_c$. We estimate the scalar precision $s$ based on the statistics of the diagonal entry (the probability of class $c$):
\begin{equation}
    s = \frac{\mu_c (1 - \mu_c)}{\sigma_c^2} - 1
\end{equation}
To ensure numerical stability, we clip $s \ge 0.1$. The parameter vector is then derived as:
\begin{equation}
    \boldsymbol{\alpha}_c = \boldsymbol{\mu} \cdot s
\end{equation}
During training, the target for an image in class $c$ is sampled as $\mathbf{y} \sim \text{Dir}(\boldsymbol{\alpha}_c)$.

\subsection{Training Methods}

\subsubsection{Loss Function}
For methods utilizing probabilistic targets (Average Soft-Labels or Dirichlet), we minimize the Kullback-Leibler (KL) Divergence. Let $\mathbf{y}_{target}$ be the soft target and $\hat{\mathbf{y}} = \log \sigma(f_{student}(\mathbf{x}))$. The loss is:
\begin{equation}
    \mathcal{L}_{KL} = \frac{1}{B} \sum_{b=1}^B \sum_{j=1}^C \mathbf{y}_{target, j}^{(b)} \cdot \left( \log \mathbf{y}_{target, j}^{(b)} - \hat{\mathbf{y}}_{j}^{(b)} \right)
\end{equation}
If importance weighting is applied, the loss becomes a weighted sum: $\mathcal{L} = \frac{1}{B} \sum_{b=1}^B w_b \cdot D_{KL}(\mathbf{y}_{target}^{(b)} || \exp(\hat{\mathbf{y}}^{(b)})).$

\subsubsection{Importance Weighting}
\label{sec:app:subsubsection:importance_weighting}
We translate uncertainty scores (e.g. energy, entropy, etc.) into importance weights to modulate the loss function. Given scores $\mathcal{S}$ for the active dataset, the weight $w_i$ for sample $i$ is computed using a Boltzmann distribution and normalized to unit mean:
\begin{equation}
    w_i' = \exp\left(-\frac{s_i}{T_{weight}}\right), \quad w_i = \frac{w_i'}{\frac{1}{N}\sum_{j} w_j'}
\end{equation}
This assigns higher weights to samples with lower uncertainty scores (lower energy/entropy).

\section{Compression Experiments Full Details}
\label{sec:app:compression}
In this section, we report the compression sizes obtained under different pruning rates. The results illustrate how the pruning rate and the number of classes in the target dataset affect the overall payload size. Each table reports the size of the raw data and the size of the compact representation (obtained by storing labels and indices using the smallest possible integer type), as well as the compressed sizes after applying Huffman coding and Zstandard (Zstd). 

In addition, we compare two representations for storing the selected indices: integer lists (idx) and binary masks (bmp). In the compact representation, we save the indices using delta encoding—storing the difference from the previous index and casting to \textit{uint8} or \textit{uint16} where possible. If filtering is not used, the payload size consists only of the hard labels for all images in the reference set.

\begin{table}[p]
\centering
\caption{ \textbf{Compression Summary for Caltech-101, CIFAR-10, CUB-200-2011.} In \textbf{bold} we highlight the best compression for every 2 rows (as we also compare between bitmap and delta indices encodings for the pruning bit). 
}
\label{tab:comp_caltech}
\vskip 0.15in
\begin{small}
\setlength{\tabcolsep}{5pt}
\begin{sc}
\begin{tabular}{l @{\hspace{8pt}} cccc @{\hspace{8pt}} cccc}
\toprule
\abovespace
& \multicolumn{4}{c}{ImageNet-21K (14.2M images)} & \multicolumn{4}{c}{ImageNet-1K (1.2M images)} \\
\cmidrule(lr){2-5} \cmidrule(lr){6-9}
$p$ & Raw  & Compact & Huffman & Zstd & Raw  & Compact & Huffman & Zstd \\
\midrule
\abovespace
0.1\% (Idx)  & 83.19 KB & 69.32 KB & 21.66 KB & \textbf{20.95 KB} & 7.51 KB & 6.25 KB & \textbf{1.57 KB} & 1.73 KB \\
0.1\% (Bmp)  & 1.72 MB & 1.71 MB & 230.84 KB & 21.61 KB & 158.89 KB & 157.64 KB & 20.71 KB & 1.62 KB \\
0.5\% (Idx)  & 415.93 KB & 346.61 KB & 106.16 KB & 95.39 KB & 37.53 KB & 18.76 KB & 8.63 KB & 6.77 KB \\
0.5\% (Bmp)  & 1.83 MB & 1.76 MB & 289.91 KB & \textbf{84.14 KB} & 168.90 KB & 162.65 KB & 26.05 KB & \textbf{6.52 KB} \\
1\% (Idx)    & 831.86 KB & 415.93 KB & 200.09 KB & 151.80 KB & 75.06 KB & 37.53 KB & 16.86 KB & 12.59 KB \\
1\% (Bmp)    & 1.96 MB & 1.83 MB & 360.14 KB & \textbf{148.91 KB} & 181.41 KB & 168.90 KB & 32.57 KB & \textbf{11.71 KB} \\
5\% (Idx)    & 4.06 MB & 2.03 MB & 810.29 KB & 583.75 KB & 375.34 KB & 187.67 KB & 72.60 KB & 50.27 KB \\
5\% (Bmp)    & 3.05 MB & 2.37 MB & 883.88 KB & \textbf{540.18 KB} & 281.51 KB & 218.95 KB & 80.67 KB & \textbf{45.44 KB} \\
10\% (Idx)   & 8.12 MB & 4.06 MB & 1.44 MB &  1.01 MB & 750.68 KB & 375.34 KB & 135.85 KB & 91.50 KB \\
10\% (Bmp)   & 4.40 MB & 3.05 MB & 1.47 MB & \textbf{941.23 KB} & 406.62 KB & 281.51 KB & 138.95 KB & \textbf{82.20 KB} \\
25\% (Idx)   & 20.31 MB & 10.15 MB & 3.28 MB & 2.22 MB & 1.83 MB & 938.35 KB & 310.56 KB & 206.70 KB \\
25\% (Bmp)   & 8.46 MB & 5.08 MB & 3.27 MB & \textbf{2.04 MB} & 781.96 KB & 469.18 KB & 309.26 KB & \textbf{186.83 KB} \\
50\% (Idx)   & 40.62 MB & 13.54 MB & 5.92 MB & 3.90 MB & 3.67 MB & 1.22 MB & 558.56 KB & 357.13 KB \\
50\% (Bmp)   & 15.23 MB & 8.46 MB & 5.87 MB &\textbf{ 3.78 MB} & 1.37 MB & 781.96 KB & 553.09 KB & \textbf{348.83 KB} \\
No Filter    & 27.08 MB & 13.54 MB & 9.26 MB & \textbf{6.15 MB} & 2.44 MB & 1.22 MB & 861.51 KB & \textbf{555.29 KB} \\
\midrule
\abovespace
0.1\% (Idx)  & 83.19 KB & 69.32 KB &\textbf{ 14.27 KB} & 16.97 KB & 7.51 KB & 6.25 KB & \textbf{1.36 KB} & 1.83 KB \\
0.1\% (Bmp)  & 1.72 MB & 1.71 MB & 225.39 KB & 16.12 KB & 158.89 KB & 157.64 KB & 20.36 KB & 1.77 KB \\
0.5\% (Idx)  & 415.93 KB & 346.61 KB & 64.99 KB & 61.34 KB & 37.53 KB & 18.76 KB & 6.08 KB & 5.85 KB \\
0.5\% (Bmp)  & 1.83 MB & 1.76 MB & 263.75 KB &\textbf{ 55.16 KB} & 168.90 KB & 162.65 KB & 23.84 KB & \textbf{5.64 KB} \\
1\% (Idx)    & 831.86 KB & 693.21 KB & 126.72 KB & 107.64 KB & 75.06 KB & 37.53 KB & 11.88 KB & 10.59 KB \\
1\% (Bmp)    & 1.96 MB & 1.83 MB & 312.04 KB & \textbf{97.01 KB} & 181.41 KB & 168.90 KB & 28.34 KB & \textbf{9.93 KB} \\
5\% (Idx)    & 4.06 MB & 2.03 MB & 630.74 KB & 440.13 KB & 375.34 KB & 187.67 KB & 59.05 KB & 45.10 KB \\
5\% (Bmp)    & 3.05 MB & 2.37 MB & 731.43 KB &\textbf{ 401.63 KB} & 281.51 KB & 218.95 KB & 68.02 KB & \textbf{39.91 KB} \\
10\% (Idx)   & 8.12 MB & 4.06 MB & 1.15 MB & 818.21 KB & 750.68 KB & 375.34 KB & 104.41 KB & 76.93 KB \\
10\% (Bmp)   & 4.40 MB & 3.05 MB & 1.18 MB & \textbf{759.21 KB} & 406.62 KB & 281.51 KB & 110.48 KB & \textbf{67.80 KB} \\
25\% (Idx)   & 20.31 MB & 10.15 MB & 2.59 MB & 1.99 MB & 1.83 MB & 938.35 KB & 224.10 KB & 166.46 KB \\
25\% (Bmp)   & 8.46 MB & 5.08 MB & 2.54 MB & \textbf{1.83 MB} & 781.96 KB & 469.18 KB & 223.66 KB & \textbf{149.08 KB} \\
50\% (Idx)   & 40.62 MB & 13.54 MB & 4.34 MB & 3.45 MB & 3.67 MB & 1.22 MB & 392.00 KB & 291.59 KB \\
50\% (Bmp)   & 15.23 MB & 8.46 MB & 4.26 MB & \textbf{3.27 MB} & 1.37 MB & 781.96 KB & 380.68 KB & \textbf{280.51 KB} \\
No Filter     & 27.08 MB & 13.54 MB & 5.35 MB &\textbf{ 4.31 MB} & 2.44 MB & 1.22 MB & 488.71 KB & \textbf{397.41 KB} \\
\midrule
\abovespace
0.1\% (Idx)  & 83.19 KB & 69.32 KB & 19.56 KB & 16.12 KB & 7.51 KB & 6.25 KB & 1.17 KB & 1.36 KB \\
0.1\% (Bmp)  & 1.72 MB & 1.71 MB & 233.01 KB & \textbf{14.07 KB} & 158.89 KB & 157.64 KB & 20.53 KB & \textbf{1.08 KB} \\
0.5\% (Idx)  & 415.93 KB & 346.61 KB & 87.61 KB & 53.37 KB & 37.53 KB & 31.27 KB & 5.12 KB & 3.94 KB \\
0.5\% (Bmp)  & 1.83 MB & 1.76 MB & 297.93 KB & \textbf{45.48 KB} & 168.90 KB & 162.65 KB & 24.42 KB & \textbf{3.15 KB} \\
1\% (Idx)    & 831.86 KB & 693.21 KB & 169.38 KB & 97.41 KB & 75.06 KB & 62.55 KB & 12.38 KB & 9.00 KB \\
1\% (Bmp)    & 1.96 MB & 1.83 MB & 375.60 KB & \textbf{84.83 KB} & 181.41 KB & 168.90 KB & 30.39 KB & \textbf{7.86 KB} \\
5\% (Idx)    & 4.06 MB & 2.03 MB & 916.85 KB & 643.00 KB & 375.34 KB & 187.67 KB & 79.67 KB & 62.05 KB \\
5\% (Bmp)    & 3.05 MB & 2.37 MB & 1.00 MB & \textbf{620.14 KB} & 281.51 KB & 218.95 KB & 89.24 KB & \textbf{58.36 KB} \\
10\% (Idx)   & 8.12 MB & 4.06 MB & 1.76 MB & 1.36 MB & 750.68 KB & 375.34 KB & 157.59 KB & 129.12 KB \\
10\% (Bmp)   & 4.40 MB & 3.05 MB & 1.78 MB & \textbf{1.30 MB} & 406.62 KB & 281.51 KB & 161.13 KB & \textbf{120.80 KB} \\
25\% (Idx)   & 20.31 MB & 10.15 MB & 4.13 MB & 3.48 MB & 1.83 MB & 938.35 KB & 372.65 KB & 314.23 KB \\
25\% (Bmp)   & 8.46 MB & 5.08 MB & 4.09 MB &\textbf{ 3.33 MB} & 781.96 KB & 469.18 KB & 370.30 KB & \textbf{299.80 KB} \\
50\% (Idx)   & 40.62 MB & 20.31 MB & 7.56 MB & 6.58 MB & 3.67 MB & 1.22 MB & 694.00 KB & 578.75 KB \\
50\% (Bmp)   & 15.23 MB & 8.46 MB & 7.43 MB & \textbf{6.34 MB} & 1.37 MB & 781.96 KB & 679.95 KB & \textbf{569.95 KB} \\
No Filter     & 27.08 MB & 13.54 MB & 12.09 MB & \textbf{10.50 MB} & 2.44 MB & 1.22 MB & 1.09 MB &\textbf{ 934.45 KB} \\

\bottomrule
\end{tabular}
\end{sc}
\end{small}
\end{table}

\begin{table}[p]
\centering
\caption{ \textbf{Compression Summary for DTD, FGVC-AIRCRAFT, FOOD-101}
}
\label{tab:comp_dtd}
\vskip 0.15in
\begin{small}
\setlength{\tabcolsep}{5pt}
\begin{sc}
\begin{tabular}{l @{\hspace{8pt}} cccc @{\hspace{8pt}} cccc}
\toprule
\abovespace
& \multicolumn{4}{c}{ImageNet-21K (14.2M images)} & \multicolumn{4}{c}{ImageNet-1K (1.2M images)} \\
\cmidrule(lr){2-5} \cmidrule(lr){6-9}
$p$ & Raw  & Compact & Huffman & Zstd & Raw  & Compact & Huffman & Zstd \\
\midrule
\abovespace

0.1\% (Idx)  & 83.19 KB & 69.32 KB & \textbf{21.10 KB} & 22.48 KB & 7.51 KB & 3.75 KB & \textbf{1.46 KB} & 1.61 KB \\
0.1\% (Bmp)  & 1.72 MB & 1.71 MB & 230.01 KB & 23.76 KB & 158.89 KB & 157.64 KB & 20.55 KB & 1.91 KB \\
0.5\% (Idx)  & 415.93 KB & 207.96 KB & 102.81 KB & \textbf{91.67 KB} & 37.53 KB & 18.76 KB & 7.94 KB & 7.50 KB \\
0.5\% (Bmp)  & 1.83 MB & 1.76 MB & 286.34 KB & 95.17 KB & 168.90 KB & 162.65 KB & 25.03 KB & \textbf{7.47 KB} \\
1\% (Idx)    & 831.86 KB & 415.93 KB & 197.48 KB & 175.57 KB & 75.06 KB & 37.53 KB & 15.83 KB & 14.57 KB \\
1\% (Bmp)    & 1.96 MB & 1.83 MB & 356.29 KB & \textbf{174.52 KB} & 181.41 KB & 168.90 KB & 30.75 KB & \textbf{13.98 KB} \\
5\% (Idx)    & 4.06 MB & 2.03 MB & 846.35 KB & 743.57 KB & 375.34 KB & 187.67 KB & 71.96 KB & 65.29 KB \\
5\% (Bmp)    & 3.05 MB & 2.37 MB & 900.91 KB & \textbf{700.65} KB & 281.51 KB & 218.95 KB & 77.12 KB & \textbf{58.62 KB} \\
10\% (Idx)   & 8.12 MB & 4.06 MB & 1.51 MB & 1.33 MB & 750.68 KB & 375.34 KB & 133.70 KB & 122.18 KB \\
10\% (Bmp)   & 4.40 MB & 3.05 MB & 1.52 MB & \textbf{1.25 MB} & 406.62 KB & 281.51 KB & 134.46 KB & \textbf{109.68 KB} \\
25\% (Idx)   & 20.31 MB & 10.15 MB & 3.29 MB & 2.92 MB & 1.83 MB & 938.35 KB & 295.75 KB & 264.04 KB \\
25\% (Bmp)   & 8.46 MB & 5.08 MB & 3.26 MB &\textbf{ 2.74 MB} & 781.96 KB & 469.18 KB & 292.71 KB & \textbf{241.75 KB} \\
50\% (Idx)   & 40.62 MB & 13.54 MB & 5.69 MB & 4.93 MB & 3.67 MB & 1.22 MB & 516.89 KB & 434.73 KB \\
50\% (Bmp)   & 15.23 MB & 8.46 MB & 5.65 MB & \textbf{4.79 MB} & 1.37 MB & 781.96 KB & 512.48 KB & \textbf{426.67 KB} \\
No Filter     & 27.08 MB & 13.54 MB & 8.42 MB &\textbf{ 7.05 MB} & 2.44 MB & 1.22 MB & 764.40 KB & \textbf{626.21 KB} \\
\midrule
\abovespace

0.1\% (Idx)  & 83.19 KB & 69.32 KB &\textbf{ 19.33 KB} & 23.38 KB & 7.51 KB & 6.25 KB & \textbf{1.75 KB} & 2.53 KB \\
0.1\% (Bmp)  & 1.72 MB & 1.71 MB & 230.82 KB & 22.95 KB & 158.89 KB & 157.64 KB & 20.80 KB & 2.55 KB \\
0.5\% (Idx)  & 415.93 KB & 346.61 KB & \textbf{95.21 KB} & 112.46 KB & 37.53 KB & 18.76 KB & \textbf{8.62 KB} & 9.69 KB \\
0.5\% (Bmp)  & 1.83 MB & 1.76 MB & 283.53 KB & 100.20 KB & 168.90 KB & 162.65 KB & 25.55 KB & 9.67 KB \\
1\% (Idx)    & 831.86 KB & 415.93 KB & \textbf{184.86 KB} & 195.50 KB & 75.06 KB & 37.53 KB & \textbf{16.57 KB} & 18.53 KB \\
1\% (Bmp)    & 1.96 MB & 1.83 MB & 350.40 KB & 189.62 KB & 181.41 KB & 168.90 KB & 31.57 KB & 17.41 KB \\
5\% (Idx)    & 4.06 MB & 2.03 MB & 833.67 KB & 853.52 KB & 375.34 KB & 187.67 KB & 74.00 KB & 79.67 KB \\
5\% (Bmp)    & 3.05 MB & 2.37 MB & 898.83 KB & \textbf{803.97 KB} & 281.51 KB & 218.95 KB & 80.20 KB & \textbf{73.73 KB} \\
10\% (Idx)   & 8.12 MB & 4.06 MB & 1.54 MB & 1.56 MB & 750.68 KB & 375.34 KB & 140.83 KB & 146.98 KB \\
10\% (Bmp)   & 4.40 MB & 3.05 MB & 1.56 MB &\textbf{ 1.49 MB} & 406.62 KB & 281.51 KB & 141.38 KB & \textbf{136.31 KB} \\
25\% (Idx)   & 20.31 MB & 10.15 MB & 3.57 MB & 3.53 MB & 1.83 MB & 938.35 KB & 327.00 KB & 332.79 KB \\
25\% (Bmp)   & 8.46 MB & 5.08 MB & 3.54 MB & \textbf{3.36 MB} & 781.96 KB & 469.18 KB & 323.50 KB & \textbf{309.20 KB} \\
50\% (Idx)   & 40.62 MB & 13.54 MB & 6.48 MB & 6.18 MB & 3.67 MB & 1.22 MB & 596.62 KB & 570.83 KB \\
50\% (Bmp)   & 15.23 MB & 8.46 MB & 6.43 MB & \textbf{6.03 MB} & 1.37 MB & 781.96 KB & 591.94 KB & \textbf{562.64 KB} \\
No Filter     & 27.08 MB & 13.54 MB & 10.19 MB & \textbf{9.35 MB} & 2.44 MB & 1.22 MB & 940.98 KB & \textbf{857.63 KB} \\
\midrule
\abovespace

0.1\% (Idx)  & 83.19 KB & 69.32 KB & 20.79 KB & 19.50 KB & 7.51 KB & 6.25 KB & \textbf{1.38 KB} & 1.67 KB \\
0.1\% (Bmp)  & 1.72 MB & 1.71 MB & 232.39 KB & \textbf{17.98 KB} & 158.89 KB & 157.64 KB & 20.62 KB & 1.53 KB \\
0.5\% (Idx)  & 415.93 KB & 346.61 KB & 97.64 KB & 79.87 KB & 37.53 KB & 18.76 KB & 7.42 KB & 6.62 KB \\
0.5\% (Bmp)  & 1.83 MB & 1.76 MB & 296.48 KB & \textbf{74.94 KB} & 168.90 KB & 162.65 KB & 25.46 KB & \textbf{6.45 KB} \\
1\% (Idx)    & 831.86 KB & 415.93 KB & 197.60 KB & 158.06 KB & 75.06 KB & 37.53 KB & 16.70 KB & 15.24 KB \\
1\% (Bmp)    & 1.96 MB & 1.83 MB & 376.51 KB & \textbf{155.19 KB} & 181.41 KB & 168.90 KB & 32.34 KB & \textbf{14.51 KB} \\
5\% (Idx)    & 4.06 MB & 2.03 MB & 942.85 KB & 840.43 KB & 375.34 KB & 187.67 KB & 83.94 KB & 82.62 KB \\
5\% (Bmp)    & 3.05 MB & 2.37 MB & 999.63 KB & \textbf{802.00 KB} & 281.51 KB & 218.95 KB & 87.62 KB & \textbf{77.31 KB} \\
10\% (Idx)   & 8.12 MB & 4.06 MB & 1.72 MB & 1.58 MB & 750.68 KB & 375.34 KB & 155.74 KB & 155.63 KB \\
10\% (Bmp)   & 4.40 MB & 3.05 MB & 1.71 MB & \textbf{1.50 MB} & 406.62 KB & 281.51 KB & 155.20 KB & \textbf{142.45 KB} \\
25\% (Idx)   & 20.31 MB & 10.15 MB & 3.80 MB & 3.57 MB & 1.83 MB & 938.35 KB & 342.20 KB & 341.08 KB \\
25\% (Bmp)   & 8.46 MB & 5.08 MB & 3.75 MB & \textbf{3.38 MB} & 781.96 KB & 469.18 KB & 339.64 KB & \textbf{314.62 KB} \\
50\% (Idx)   & 40.62 MB & 13.54 MB & 6.64 MB & 6.18 MB & 3.67 MB & 1.22 MB & 599.40 KB & 582.58 KB \\
50\% (Bmp)   & 15.23 MB & 8.46 MB & 6.61 MB & \textbf{6.05 MB} & 1.37 MB & 781.96 KB & 597.40 KB & \textbf{562.63 KB} \\
No Filter     & 27.08 MB & 13.54 MB & 10.19 MB &\textbf{ 9.38 MB} & 2.44 MB & 1.22 MB & 924.77 KB & \textbf{871.44 KB} \\

\bottomrule
\end{tabular}
\end{sc}
\end{small}
\end{table}

\begin{table}[p]
\centering
\caption{\textbf{Compression Summary for Oxford-Flowers102, Oxford-IIIT-Pet, Places365.} Note that compression is less effective on Places365 due to its larger number of classes and reduced redundancy.}
\label{tab:comp_oxford_flowers102}
\vskip 0.15in
\begin{small}
\setlength{\tabcolsep}{5pt}
\begin{sc}
\begin{tabular}{l @{\hspace{8pt}} cccc @{\hspace{8pt}} cccc}
\toprule
\abovespace
& \multicolumn{4}{c}{ImageNet-21K (14.2M images)} & \multicolumn{4}{c}{ImageNet-1K (1.2M images)} \\
\cmidrule(lr){2-5} \cmidrule(lr){6-9}
$p$ & Raw  & Compact & Huffman & Zstd & Raw  & Compact & Huffman & Zstd \\
\midrule
\abovespace

0.1\% (Idx) & 83.19 KB & 69.32 KB & 18.69 KB & 15.95 KB & 7.51 KB & 6.25 KB & \textbf{0.96 KB} & 1.32 KB \\
0.1\% (Bmp) & 1.72 MB & 1.71 MB & 231.20 KB & \textbf{14.31 KB} & 158.89 KB & 157.64 KB & 20.37 KB & 1.13 KB \\
0.5\% (Idx) & 415.93 KB & 346.61 KB & 87.85 KB & 60.86 KB & 37.53 KB & 18.76 KB & 7.23 KB & 6.80 KB \\
0.5\% (Bmp) & 1.83 MB & 1.76 MB & 292.83 KB &\textbf{ 51.47 KB} & 168.90 KB & 162.65 KB & 24.94 KB & \textbf{7.07 KB} \\
1\% (Idx) & 831.86 KB & 693.21 KB & 168.27 KB & 107.58 KB & 75.06 KB & 37.53 KB & 16.24 KB & 14.81 KB \\
1\% (Bmp) & 1.96 MB & 1.83 MB & 366.74 KB & \textbf{90.95 KB} & 181.41 KB & 168.90 KB & 31.97 KB & \textbf{14.31 KB} \\
5\% (Idx) & 4.06 MB & 2.03 MB & 801.74 KB & 585.52 KB & 375.34 KB & 187.67 KB & 82.33 KB & 78.01 KB \\
5\% (Bmp) & 3.05 MB & 2.37 MB & 942.21 KB &\textbf{ 558.34 KB} & 281.51 KB & 218.95 KB & 88.58 KB & \textbf{72.10 KB} \\
10\% (Idx) & 8.12 MB & 4.06 MB & 1.53 MB & 1.19 MB & 750.68 KB & 375.34 KB & 156.76 KB & 145.79 KB \\
10\% (Bmp) & 4.40 MB & 3.05 MB & 1.60 MB & \textbf{1.15 MB} & 406.62 KB & 281.51 KB & 157.31 KB & \textbf{135.94 KB} \\
25\% (Idx) & 20.31 MB & 10.15 MB & 3.70 MB & 3.12 MB & 1.83 MB & 938.35 KB & 355.21 KB & 325.23 KB \\
25\% (Bmp) & 8.46 MB & 5.08 MB & 3.67 MB & \textbf{2.98 MB} & 781.96 KB & 469.18 KB & 353.43 KB & \textbf{305.19 KB} \\
50\% (Idx) & 40.62 MB & 20.31 MB & 6.83 MB & 5.99 MB & 3.67 MB & 1.22 MB & 638.19 KB & 551.53 KB \\
50\% (Bmp) & 15.23 MB & 8.46 MB & 6.72 MB & \textbf{5.74 MB} & 1.37 MB & 781.96 KB & 629.88 KB & \textbf{533.90 KB} \\
No Filter & 27.08 MB & 13.54 MB & 10.66 MB & \textbf{9.09 MB} & 2.44 MB & 1.22 MB & 1000.91 KB & \textbf{787.84 KB} \\
\midrule
\abovespace

0.1\% (Idx) & 83.19 KB & 69.32 KB & 17.85 KB & 16.54 KB & 7.51 KB & 6.25 KB & \textbf{1.62 KB} & 1.94 KB \\
0.1\% (Bmp) & 1.72 MB & 1.71 MB & 230.72 KB & \textbf{14.98 KB} & 158.89 KB & 157.64 KB & 20.66 KB & 1.74 KB \\
0.5\% (Idx) & 415.93 KB & 346.61 KB & 78.89 KB & 64.69 KB & 37.53 KB & 31.27 KB & 7.47 KB & 6.46 KB \\
0.5\% (Bmp) & 1.83 MB & 1.76 MB & 284.03 KB & \textbf{57.51 KB} & 168.90 KB & 162.65 KB & 25.62 KB & \textbf{5.41 KB} \\
1\% (Idx) & 831.86 KB & 415.93 KB & 153.19 KB & 119.06 KB & 75.06 KB & 62.55 KB & 14.02 KB & 10.83 KB \\
1\% (Bmp) & 1.96 MB & 1.83 MB & 347.45 KB & \textbf{115.40 KB} & 181.41 KB & 168.90 KB & 31.68 KB & \textbf{8.84 KB} \\
5\% (Idx) & 4.06 MB & 2.03 MB & 777.64 KB & 690.47 KB & 375.34 KB & 187.67 KB & 58.95 KB & 39.16 KB \\
5\% (Bmp) & 3.05 MB & 2.37 MB & 851.69 KB & \textbf{658.45 KB} & 281.51 KB & 218.95 KB & 74.57 KB & \textbf{35.13 KB} \\
10\% (Idx) & 8.12 MB & 4.06 MB & 1.50 MB & 1.36 MB & 750.68 KB & 375.34 KB & 108.29 KB & 75.07 KB \\
10\% (Bmp) & 4.40 MB & 3.05 MB & 1.48 MB & \textbf{1.30 MB} & 406.62 KB & 281.51 KB & 120.47 KB & \textbf{68.14 KB} \\
25\% (Idx) & 20.31 MB & 10.15 MB & 3.34 MB & 3.11 MB & 1.83 MB & 938.35 KB & 290.74 KB & 221.75 KB \\
25\% (Bmp) & 8.46 MB & 5.08 MB & 3.29 MB & \textbf{2.94 MB} & 781.96 KB & 469.18 KB & 280.72 KB & \textbf{210.13 KB} \\
50\% (Idx) & 40.62 MB & 13.54 MB & 5.75 MB & 5.27 MB & 3.67 MB & 1.22 MB & 536.90 KB & 426.33 KB \\
50\% (Bmp) & 15.23 MB & 8.46 MB & 5.71 MB & \textbf{5.12 MB} & 1.37 MB & 781.96 KB & 521.38 KB & \textbf{422.65 KB} \\
No Filter & 27.08 MB & 13.54 MB & 8.31 MB & \textbf{7.38 MB} & 2.44 MB & 1.22 MB & 779.40 KB & \textbf{626.22 KB} \\
\midrule
\abovespace

0.1\% (Idx) & 83.19 KB & 83.19 KB & \textbf{20.99 KB} & 21.11 KB & 7.51 KB & 7.51 KB & \textbf{1.82 KB} & 2.27 KB \\
0.1\% (Bmp) & 1.72 MB & 1.72 MB & 231.18 KB & 21.80 KB & 158.89 KB & 158.89 KB & 20.80 KB & 2.29 KB \\
0.5\% (Idx) & 415.93 KB & 277.29 KB & 107.31 KB & \textbf{89.19 KB }& 37.53 KB & 25.02 KB & 8.94 KB & 8.23 KB \\
0.5\% (Bmp) & 1.83 MB & 1.83 MB & 296.30 KB & 91.05 KB & 168.90 KB & 168.90 KB & 26.28 KB & \textbf{8.12 KB }\\
1\% (Idx) & 831.86 KB & 554.57 KB & 217.09 KB & 183.05 KB & 75.06 KB & 50.04 KB & 18.19 KB & 16.12 KB \\
1\% (Bmp) & 1.96 MB & 1.96 MB & 381.59 KB & \textbf{177.56 KB} & 181.41 KB & 181.41 KB & 33.70 KB & \textbf{15.24 KB} \\
5\% (Idx) & 4.06 MB & 2.71 MB & 995.15 KB & 820.58 KB & 375.34 KB & 250.23 KB & 91.88 KB & 74.87 KB \\
5\% (Bmp) & 3.05 MB & 3.05 MB & 1.05 MB & \textbf{778.06 KB} & 281.51 KB & 281.51 KB & 98.31 KB & \textbf{71.98 KB} \\
10\% (Idx) & 8.12 MB & 5.42 MB & 1.83 MB & 1.49 MB & 750.68 KB & 500.45 KB & 177.37 KB & 142.36 KB \\
10\% (Bmp) & 4.40 MB & 4.40 MB & 1.87 MB & \textbf{1.41 MB} & 406.62 KB & 406.62 KB & 179.47 KB & \textbf{136.92 KB} \\
25\% (Idx) & 20.31 MB & 13.54 MB & 4.24 MB & 3.36 MB & 1.83 MB & 1.22 MB & 405.40 KB & 328.15 KB \\
25\% (Bmp) & 8.46 MB & 8.46 MB & 4.23 MB & \textbf{3.23 MB} & 781.96 KB & 781.96 KB & 403.47 KB & \textbf{308.28 KB} \\
50\% (Idx) & 40.62 MB & 20.31 MB & 7.87 MB & 6.02 MB & 3.67 MB & 1.83 MB & 728.51 KB & 591.22 KB \\
50\% (Bmp) & 15.23 MB & 15.23 MB & 7.77 MB & \textbf{5.94 MB} & 1.37 MB & 1.37 MB & 723.03 KB & \textbf{584.04 KB} \\
No Filter & 27.08 MB & 27.08 MB & 12.83 MB & \textbf{10.50 MB} & 2.44 MB & 2.44 MB & 1.14 MB & \textbf{1020.62 KB} \\

\bottomrule
\end{tabular}
\end{sc}
\end{small}
\end{table}

\begin{table}[p]
\centering
\caption{ \textbf{Compression Summary for RESISC45}
}
\label{tab:comp_resisc45}
\vskip 0.15in
\begin{small}
\setlength{\tabcolsep}{5pt}
\begin{sc}
\begin{tabular}{l @{\hspace{8pt}} cccc @{\hspace{8pt}} cccc}
\toprule
\abovespace
& \multicolumn{4}{c}{ImageNet-21K (14.2M images)} & \multicolumn{4}{c}{ImageNet-1K (1.2M images)} \\
\cmidrule(lr){2-5} \cmidrule(lr){6-9}
$p$ & Raw  & Compact & Huffman & Zstd & Raw  & Compact & Huffman & Zstd \\
\midrule
\abovespace

0.1\% (Idx) & 83.19 KB & 41.59 KB & \textbf{19.83 KB} & 22.82 KB & 7.51 KB & 3.75 KB & \textbf{1.79 KB} & 2.35 KB \\
0.1\% (Bmp) & 1.72 MB & 1.71 MB & 226.58 KB & 25.10 KB & 158.89 KB & 157.64 KB & 20.44 KB & 3.03 KB \\
0.5\% (Idx) & 415.93 KB & 207.96 KB & \textbf{87.16 KB} & 94.19 KB & 37.53 KB & 18.76 KB & \textbf{8.58 KB} & 9.92 KB \\
0.5\% (Bmp) & 1.83 MB & 1.76 MB & 270.34 KB & 91.43 KB & 168.90 KB & 162.65 KB & 24.27 KB & 10.31 KB \\
1\% (Idx) & 831.86 KB & 415.93 KB & \textbf{162.79 KB} & 172.20 KB & 75.06 KB & 37.53 KB & \textbf{16.18 KB} & 18.20 KB \\
1\% (Bmp) & 1.96 MB & 1.83 MB & 328.04 KB & 168.22 KB & 181.41 KB & 168.90 KB & 29.27 KB & 17.95 KB \\
5\% (Idx) & 4.06 MB & 2.03 MB & 668.71 KB & 689.10 KB & 375.34 KB & 187.67 KB & 68.82 KB & 72.35 KB \\
5\% (Bmp) & 3.05 MB & 2.37 MB & 799.48 KB & \textbf{643.86 KB} & 281.51 KB & 218.95 KB & 71.90 KB & \textbf{68.78 KB} \\
10\% (Idx) & 8.12 MB & 2.71 MB & 1.16 MB & 1.11 MB & 750.68 KB & 375.34 KB & 125.58 KB & 131.29 KB \\
10\% (Bmp) & 4.40 MB & 3.05 MB & 1.32 MB & \textbf{1.10 MB} & 406.62 KB & 281.51 KB & 125.71 KB & \textbf{119.15 KB} \\
25\% (Idx) & 20.31 MB & 6.77 MB & 2.33 MB & 2.26 MB & 1.83 MB & 625.57 KB & 268.46 KB & \textbf{256.22 KB} \\
25\% (Bmp) & 8.46 MB & 5.08 MB & 2.50 MB & \textbf{2.24 MB} & 781.96 KB & 469.18 KB & 267.25 KB & 256.71 KB \\
50\% (Idx) & 40.62 MB & 33.85 MB & 4.13 MB & \textbf{2.65 MB} & 3.67 MB & 1.22 MB & 455.29 KB & 447.08 KB \\
50\% (Bmp) & 15.23 MB & 8.46 MB & 3.60 MB & 2.69 MB & 1.37 MB & 781.96 KB & 454.43 KB & \textbf{443.60 KB} \\
No Filter & 27.08 MB & 13.54 MB & 4.32 MB & 2.79 MB & 2.44 MB & 1.22 MB & 648.31 KB & \textbf{620.88 KB} \\
\bottomrule
\end{tabular}
\end{sc}
\end{small}
\end{table}

\begin{table*}[ht]
\centering
\caption{ \textbf{Compression Summary Over All Datasets: ImageNet-1K (1.2M images).} This table summarizes the compression results over all of the datasets for all $p$ values.  The values in the table are min-max sizes.
}
\label{tab:comp_sum_1k}
\vskip 0.15in
\begin{sc}
\begin{tabular*}{\textwidth}{@{\extracolsep{\fill}} l cccc}
\toprule
$p$ & Raw  & Compact & Huffman & Zstd \\
\midrule
0.1\%  & 7.51--158.89 KB & 3.75--157.64 KB & 1.46--20.71 KB & 1.08--2.53 KB \\
0.5\%  & 37.53--168.90 KB & 18.76--168.90 KB & 7.23--26.28 KB & 3.15--9.92 KB \\
1\%    & 181.41 KB & 168.90--181.41 KB & 28.34--33.70 KB & 7.86--17.95 KB \\
5\%    & 281.51 KB & 218.95--281.51 KB & 68.02--98.31 KB & 35.13--77.31 KB \\
10\%   & 406.62 KB & 281.51--406.62 KB & 110.48--179.47 KB & 67.80--142.45 KB \\
25\%   & 0.76--1.83 MB & 469.18--781.96 KB & 223.66--403.47 KB & 149.08--314.62 KB \\
50\%   & 1.37 MB & 0.76--1.37 MB & 380.68--723.03 KB & 280.51--584.04 KB \\
100\%  & 2.44 MB & 1.22--2.44 MB & 0.48--1.14 MB & 397.41--1020.62 KB \\
\bottomrule
\end{tabular*}
\end{sc}
\end{table*}

\begin{table*}[ht]
\centering
\caption{ \textbf{Compression Summary Over All Datasets: ImageNet-21K (14.2M images).} In comparison to \cref{tab:comp_sum_1k}, the payloads are roughly 10-12x larger.
}
\label{tab:comp_sum_21k}
\vskip 0.15in
\begin{sc}
\begin{tabular*}{\textwidth}{@{\extracolsep{\fill}} l cccc}
\toprule
$p$ & Raw~ & Compact & Huffman & Zstd \\
\midrule
0.1\%~ & 0.08--1.72 MB & 0.04--1.71 MB & 17.43--233.01 KB & 14.07--27.13 KB \\
0.5\%~ & 0.41--1.83 MB & 0.20--1.83 MB & 77.75--305.21 KB & 45.48--108.84 KB \\
1\%~ ~ & 0.81--1.96 MB & 0.41--1.96 MB & 151.00--396.44 KB & 84.83--206.08 KB \\
5\%~ ~ & 3.05 MB & 2.37--3.05 MB & 0.57--1.10 MB & 401.63--877.05 KB \\
10\%~ ~& 4.40--8.12 MB & 2.71--4.40 MB & 0.88--1.95 MB & 0.67--1.58 MB \\
25\%~ ~& 8.46 MB & 5.08--8.46 MB & 1.65--4.34 MB & 1.21--3.47 MB \\
50\%~ ~& 15.23--40.62 MB & 8.46--33.85 MB & 2.49--7.88 MB & 1.87--6.42 MB \\
100\%~ & 27.08 MB & 13.54--27.08 MB & 2.29--12.83 MB & 1.77--10.50 MB \\
\bottomrule
\end{tabular*}
\end{sc}
\end{table*}


\end{document}